\pdfoutput=1

\documentclass[11pt]{article}

\usepackage[final]{acl}

\usepackage{times}
\usepackage{latexsym}

\usepackage[T1]{fontenc}

\usepackage[utf8]{inputenc}

\usepackage{microtype}

\usepackage{inconsolata}

\usepackage{graphicx}

\usepackage{hyperref}
\usepackage{url}
\usepackage{booktabs}
\usepackage{bbding}
\usepackage{wrapfig}
\usepackage{multirow}
\usepackage{makecell}
\usepackage{pifont}
\usepackage{enumitem}
\usepackage{colortbl}
\usepackage{amsmath}
\usepackage{tikz}
\usepackage{array}
\usepackage{amsfonts}
\usepackage{tcolorbox}
\usepackage{subcaption}
\definecolor{BoxBackground}{RGB}{240, 240, 240}
\definecolor{BoxFrame}{RGB}{0, 0, 0}
\definecolor{TitleBackground}{RGB}{0, 0, 0}
\definecolor{TitleText}{RGB}{255, 255, 255}

\tcbset{
  academicbox/.style={
    boxsep=5pt,
    left=2pt,
    right=2pt,
    bottom=0.5pt,
    boxrule=0.5pt,
    colback=BoxBackground,
    colframe=BoxFrame,
    colbacktitle=TitleBackground,
    coltitle=TitleText,
    title={#1},
  }
}
\newtcolorbox{AcademicBox}[1][]{
    academicbox=#1
}
\definecolor{SoftBlue}{RGB}{135, 206, 250}
\definecolor{SoftOrange}{RGB}{255, 224, 178}
\definecolor{SoftGreen}{RGB}{144, 238, 144}
\definecolor{CorrectGreen}{RGB}{76, 175, 80}
\definecolor{ErrorRed}{RGB}{211, 47, 47}

\definecolor{TableRowColor}{HTML}{F3E6FA}
\definecolor{TextBackColor}{HTML}{D0DFE5}
\definecolor{gpt35}{HTML}{708090}
\definecolor{gpt4}{HTML}{4682B4}
\definecolor{gpt4turbo}{HTML}{2E8B57}
\definecolor{gpt4o}{HTML}{6A5ACD}

\newcommand{\gptthreepointfivelogo}{
\begin{tikzpicture}
    \foreach \i in {0,60,120,180,240,300} {
        \draw[gpt35,semithick,rotate=\i] (0,0.04) -- (0.06928,0) -- (0.06928,-0.08) 
        arc[start angle=330,end angle=240,radius=0.064];
    }
\end{tikzpicture}
}
\newcommand{\gptthreepointfivelogowithtext}{
\gptthreepointfivelogo GPT-3.5
}

\newcommand{\gptfourlogo}{
\begin{tikzpicture}
    \foreach \i in {0,60,120,180,240,300} {
        \draw[gpt4,semithick,rotate=\i] (0,0.04) -- (0.06928,0) -- (0.06928,-0.08) 
        arc[start angle=330,end angle=240,radius=0.064];
    }
\end{tikzpicture}
}
\newcommand{\gptfourlogowithtext}{
\gptfourlogo GPT-4
}

\newcommand{\gptfourturbologo}{
\begin{tikzpicture}
    \foreach \i in {0,60,120,180,240,300} {
        \draw[gpt4turbo,semithick,rotate=\i] (0,0.04) -- (0.06928,0) -- (0.06928,-0.08) 
        arc[start angle=330,end angle=240,radius=0.064];
    }
\end{tikzpicture}
}

\newcommand{\gptfourologo}{
\begin{tikzpicture}
    \foreach \i in {0,60,120,180,240,300} {
        \draw[gpt4o,semithick,rotate=\i] (0,0.04) -- (0.06928,0) -- (0.06928,-0.08) 
        arc[start angle=330,end angle=240,radius=0.064];
    }
\end{tikzpicture}
}
\newcommand{\gptfourologowithtext}{
\gptfourologo GPT-4o
}

\newcommand{\deepseeklogo}{
\includegraphics[scale=0.04]{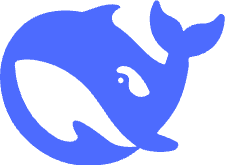}
}
\newcommand{\deepseeklogowithtext}{
\deepseeklogo DSMath-7B-RL
}

\newcommand{\qwentwologo}{
\includegraphics[scale=0.013]{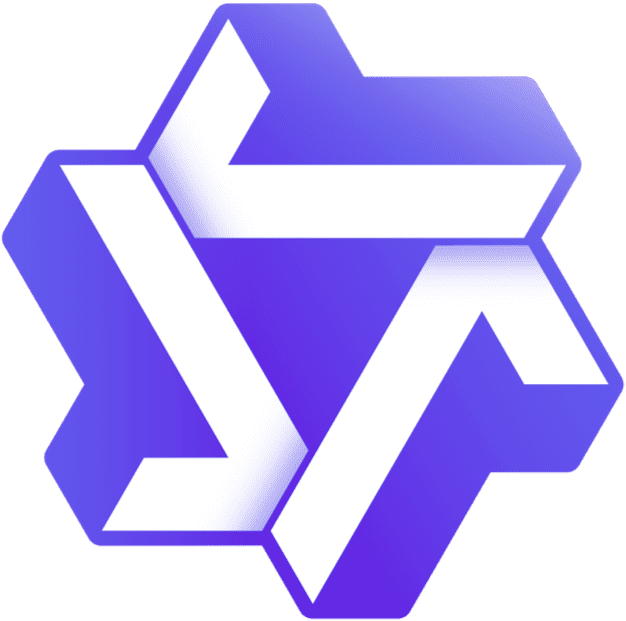}
}
\newcommand{\qwentwologowithtext}{
\qwentwologo Qwen2-Math-7B-Ins
}

\title{Unleashing LLM Reasoning Capability via Scalable Question Synthesis from Scratch}

\author{Yuyang Ding$^1$, Xinyu Shi$^1$, Xiaobo Liang$^1$, Juntao Li$^1$\thanks{\;Corresponding author} \\
\textbf{Zhaopeng Tu$^2$, Qiaoming Zhu$^1$, Min Zhang$^1$} \\
$^1$Soochow University \; $^2$Tencent AI Lab \\
\texttt{\{yyding23,xyshi02\}@stu.suda.edu.cn} \\
\texttt{\{xbliang,ljt,qmzhu,minzhang\}@suda.edu.cn} \quad \texttt{zptu@tencent.com}
}

\begin{document}
\maketitle
\begin{abstract}
Improving the mathematical reasoning capabilities of Large Language Models (LLMs) is critical for advancing artificial intelligence.
However, access to extensive, diverse, and high-quality reasoning datasets remains a significant challenge, particularly for the open-source community.
In this paper, we propose ScaleQuest, a novel, scalable, and cost-effective data synthesis method that enables the generation of large-scale mathematical reasoning datasets using lightweight 7B-scale models.
ScaleQuest introduces a two-stage question-tuning process comprising Question Fine-Tuning (QFT) and Question Preference Optimization (QPO) to unlock the question generation capabilities of problem-solving models.
By generating diverse questions from scratch -- without relying on powerful proprietary models or seed data -- we produce a dataset of 1 million problem-solution pairs.
Our experiments demonstrate that models trained on our data outperform existing open-source datasets in both in-domain and out-of-domain evaluations.
Furthermore, our approach shows continued performance improvement as the volume of training data increases, highlighting its potential for ongoing data scaling.
The extensive improvements observed in code reasoning tasks demonstrate the generalization capabilities of our proposed method.
Our work provides the open-source community with a practical solution to enhance the mathematical reasoning abilities of LLMs.\footnote{Project Page: \url{https://scalequest.github.io}.}
\end{abstract}

\section{Introduction}
\label{sec:introduction}

How to improve the mathematical reasoning capabilities of Large Language Models (LLMs) has attracted significant attention.
The success of recent advanced models, such as OpenAI o1 and Claude-3.5, heavily depends on access to extensive, diverse, and high-quality reasoning datasets.
However, the proprietary nature of the data presents a significant barrier to the open-source community.
Recent works have highlighted data synthesis as a promising approach~\citep{ntoutsi2020bias} to address data scarcity for instruction tuning~\citep{inan2023llama}.
As recent works have disclosed that crafting the right questions is crucial for eliciting the reasoning capabilities of LLMs~\citep{yu2023metamath,shah2024ai}, the core of reasoning data synthesis lies in creating large-scale and novel questions. 

\begin{figure}[t]
    \centering
    \includegraphics[width=1.0\linewidth]{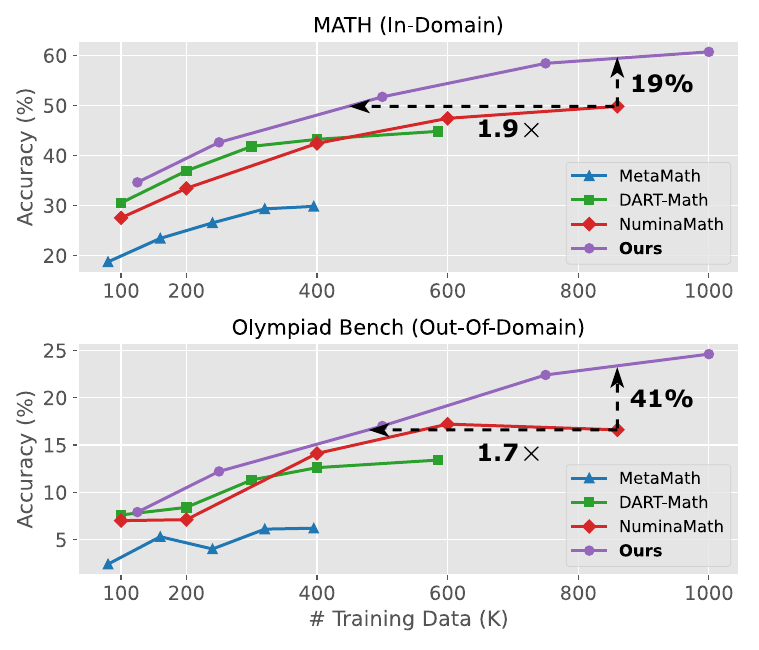}
    \caption{Results of Llama3-8B-Base fine-tuned on publicly available datasets in MATH and OlympiadBench. Our approach demonstrates strong scalability and significant potential for further improvement.}
    \label{fig:abstract_results}
\end{figure}

Previous efforts in reasoning data synthesis have demonstrated the effectiveness of leveraging powerful language models to generate instructions.
We categorize these approaches into two types: question-driven approaches and knowledge-driven approaches.
Question-driven methods include question rephrasing~\citep{yu2023metamath}, evol-instruct~\citep{xu2023wizardlm,luo2023wizardmath,zeng2024automatic}, question back-translation~\citep{lu2024mathgenie}, or providing few-shot examples~\citep{mitra2024orca}.
These methods are limited in data diversity, as the generated problems closely resemble the seed questions, with only minor modifications, such as added conditions or numerical changes.
This lack of diversity hampers their potential for scalability.
To improve question diversity, recent knowledge-driven works~\citep{huang2024mustard} scale question synthesis by constructing knowledge bases~\citep{li2024synthetic} or concept graphs~\citep{tang2024mathscale} and sampling key points~\citep{huang2024key} from them to generate new questions.
Nevertheless, the above two types of approaches rely on strong models, like GPT-4, to synthesize new questions, but the high API costs make it impractical to generate large-scale data.
Despite these advancements, the open-source community still lacks high-quality data at scale and cost-effective synthesis methods.


To meet this requirement, we explore a scalable, low-cost method for data synthesis.
We observe that using problem-solving models to directly synthesize reasoning questions, as explored in \citet{yu2023bag} and \citet{xu2024magpie}, falls short in synthesizing reasoning data, as shown in Figure~\ref{fig:abstract_results} (see Llama3-8B-Magpie results).
Accordingly, we propose a novel, scalable, and cost-effective data synthesis method, ScaleQuest, which first introduces a two-stage question-tuning process consisting of Question Fine-Tuning (QFT) and Question Preference Optimization (QPO) to unlock the question generation capability of problem-solving models.  
Once fine-tuned, these models can then generate diverse questions by sampling from a broad search space without the need for additional seed questions or knowledge constraints.
The generated questions can be further refined through a filtering process, focusing on language clarity, solvability, and appropriate difficulty.
Moreover, we introduce an extra reward-based filtering strategy to select high-quality responses.

We generate data using lightweight 7B-scale models, producing a final dataset of 1 million question-answer pairs.
As shown in Figure~\ref{fig:abstract_results}, compared with other publicly available datasets such as MetaMath~\citep{yu2023metamath}, DART-Math~\citep{tong2024dart}, and NuminaMath~\citep{numina_math_datasets}, our approach demonstrates great scalability in both in-domain and out-of-domain evaluation.
In terms of in-domain evaluation, our method outperforms existing high-quality open-source datasets, achieving better results with the same amount of data.
For out-of-domain evaluation, the performance of our approach continues to show promising trends as the volume of training data increases, indicating significant potential for further improvements through ongoing data scaling.
Additionally, we also extend our approach to long chain-of-thought and code reasoning tasks, demonstrating its effectiveness, with details in Appendix~\ref{sec:scalequest_code}.

\section{ScaleQuest: Scaling Question Synthesis from Scratch}
\label{sec:method}

\begin{figure*}[ht]
    \centering
    \includegraphics[width=1.0\linewidth]{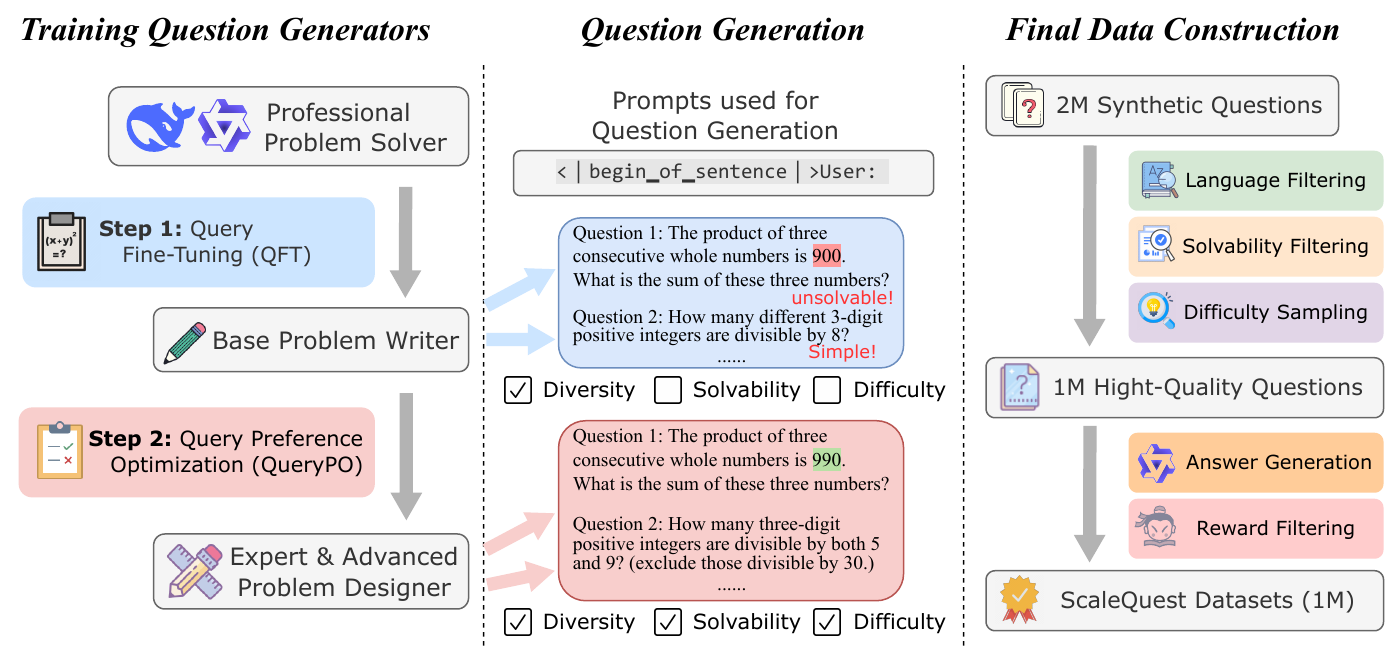}
    \caption{Overview of our ScaleQuest method. 
    }
    \label{fig:main_method}
\end{figure*}


\subsection{Question Generation from Scratch}
\label{subsec:question_generation_from_scratch}

The question generation process involves providing only a few prefix tokens from an instruction template (e.g., ``\texttt{<|begin\_of\_sentence|>User:}'') to guide the model in question generation.
A fine-tuned causal language model, which has learned to generate responses based on question-answer pairs (e.g., ``\texttt{<|begin\_of\_sentence|>User: \{Question\}. Assistant: \{Response\}}''), could potentially be leveraged to generate questions directly~\citep{xu2024magpie}.
This is because, during instruction tuning, the model is trained using a causal mask, where each token only attends to preceding tokens.
This ensures that the hidden states evolve based on past context without future token influence.
However, during instruction tuning, the actual loss is calculated based on the response, i.e.,
\begin{equation}
    \mathcal{L} = - \log P(y_i | X, y_{<i}),
\end{equation}
where $X = \{x_1, x_2, \ldots, x_m\}$ denotes question and $Y = \{y_1, y_2, \ldots, y_n\}$ denotes response.
Since $P(x_i | x_{<i})$ is inherently modeled, we need to activate the model’s capability for question generation.

\subsection{Question Fine-Tuning (QFT)}
\label{subsec:qft}


To activate the model’s question generation capability, we first perform Question Fine-Tuning (QFT), where we train the problem-solving model using a small set of problems.
To ensure that the generator stops after producing the questions and does not continue generating a response, we add an end-of-sentence token at the end of each question.
We use approximately 15K problems (without solutions) by mixing the training set of GSM8K~\citep{cobbe2021training} and MATH~\citep{hendrycks2021measuring} datasets as training samples.

\begin{figure}[t]
    \centering
    \includegraphics[width=0.9\linewidth]{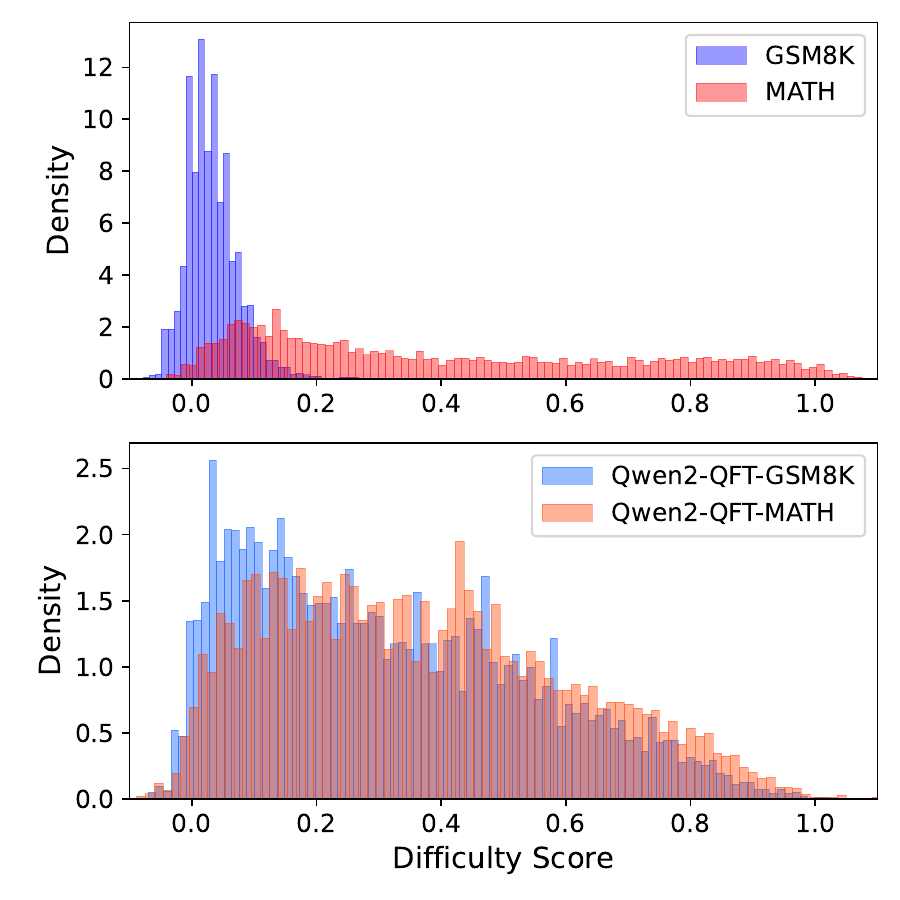}
    \caption{The difficulty distribution of two real-world datasets and two synthetic datasets. The difficulty score is calculated based solely on the problem part.}
    \label{fig:difficulty_distribution}
\end{figure}

The purpose of utilizing these problems is to activate the model’s question-generation capability rather than to make the model memorize them.
To validate this hypothesis, we train the model separately using the GSM8K and MATH datasets and compare whether the distribution of the generated questions matched that of the training data.
To evaluate the question distribution, we use a difficulty classifier, which maps a question into a difficulty score (details in Section~\ref{subsec:question_filtering}).
We perform QFT based on Qwen2-Math-7B-Instruct~\citep{yang2024qwen2}, then use the two QFT models, \texttt{Qwen2-QFT-GSM8K} and \texttt{Qwen2-QFT-MATH}, to synthesize 10K questions.
The difficulty distribution of these four datasets is shown in Figure~\ref{fig:difficulty_distribution}.
We find that the generated questions separately differed from both GSM8K and MATH, yet they both converged toward the same distribution.
This suggests that the QFT process enhances the model’s question-generation capabilities without leading to overfitting the training data.

\subsection{Question Preference Optimization (QPO)}
\label{subsec:qpo}


The model is able to generate meaningful and diverse questions after QFT, but the quality is still not high enough, as shown in Figure~\ref{fig:main_method}. 
This is reflected in two aspects: (1) solvability: the math problem should have appropriate constraints and correct answers, and (2) difficulty: the model needs to learn from more challenging problems, yet some of the generated questions are still too simple.
To address these two aspects, we apply Question Preference Optimization (QPO).

We first use the model after QFT to generate 10K questions.
Subsequently, we aim to refine these samples with a primary focus on solvability and difficulty.
We leverage an external LLM for optimization as an alternative to manual annotation.
However, we find that simultaneously optimizing both poses a challenge for the LLMs.
Therefore, for each sample, we randomly select one of the two optimization directions, prioritizing either solvability or difficulty (with optimization prompts in Figure~\ref{fig:prompt_solvability} and \ref{fig:prompt_difficulty}).
The optimized questions, denoted as  $y_w$, are treated as preferred data, while the original questions, denoted as $y_l$, are considered dispreferred data.
Inspired by Direct Preference Optimization (DPO)~\citep{rafailov2024direct}, we propose QPO for question optimization:
\begin{figure}[t]
    \centering
    \includegraphics[width=0.9\linewidth]{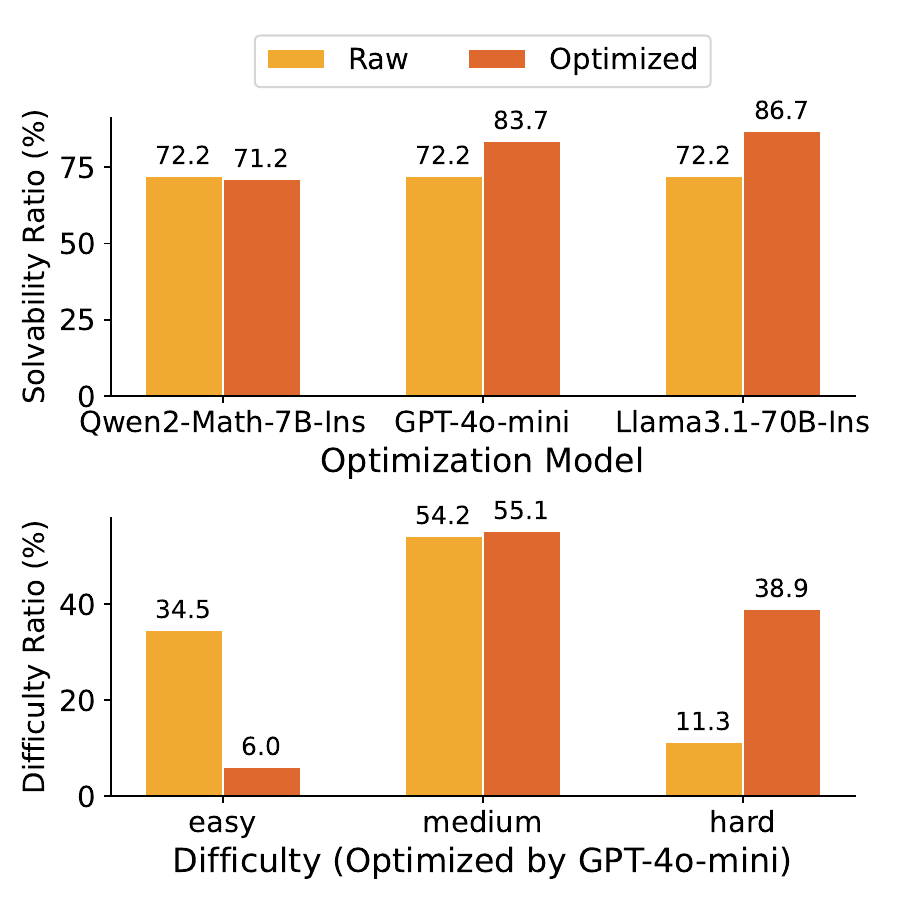}
    \caption{The solvability and difficulty of the raw questions generated by the QFT model and the optimized ones. We use GPT-4o as a manual substitute to evaluate the effectiveness of solvability and difficulty optimization, with evaluation prompt in Figure~\ref{fig:prompt_solvability_check} and \ref{fig:prompt_difficulty_check}.}
    \label{fig:solvability_and_difficulty}
\end{figure}
\begin{equation}
\label{eq:qpo}
\begin{aligned}
    & \mathcal{L}_\text{QPO}(\pi_{\theta}; \pi_\text{ref}) = 
    -\mathbb{E}_{(y_w, y_l)\sim \mathcal{D}} \bigg[ \\
    & \hspace{0.3cm} \log \sigma \bigg( 
    \beta \log \frac{\pi_{\theta}(y_w)}{\pi_\text{ref}(y_w)} 
    - \beta \log \frac{\pi_{\theta}(y_l)}{\pi_\text{ref}(y_l)}
    \bigg) \bigg].
\end{aligned}
\end{equation}



In terms of model selection, we experiment with three models: Qwen2-Math-7B-Instruct, GPT-4o-mini, and Llama3.1-70B-Instruct.
The results are shown in Figure~\ref{fig:solvability_and_difficulty}.
In terms of solvability, Qwen2-Math-7B-Instruct proves inadequate for this task, as the optimized questions result in decreased solvability.
A possible reason for this is the model’s insufficient ability to follow instructions accurately, resulting in many answers that fail to meet the specified optimization constraints.

\subsection{Question Filtering}
\label{subsec:question_filtering}

After the QFT and QPO phases, we obtain the question generators Qwen2-Math-QGen.
There are still some minor issues in the generated questions, primarily related to language, solvability, and difficulty.
To address these challenges, we apply the filtering steps:

\paragraph{Language Filtering}
The question generator models still produce a substantial number of math questions in other languages, accounting for approximately 20\%.
Since our focus is on English math questions, we remove non-English questions by identifying questions containing non-English characters and filtering out those samples.

\paragraph{Solvability Filtering}
Although QPO effectively enhances the solvability of generated questions, some questions remain nonsensical.
To filter out such samples, we use Qwen2-Math-7B-Instruct to evaluate whether the question is meaningful and whether the conditions are sufficient, with filtering prompts provided in Figure~\ref{fig:prompt_solvability_check}.

\paragraph{Difficulty Sampling}
We measure the difficulty of a question using the fail rate~\citep{tong2024dart} — the proportion of incorrect responses when sampling $n$ responses for a given question.
This metric aligns with the intuition that harder questions tend to result in fewer correct responses.
Following \citet{tong2024dart}, we use DeepSeekMath-7B-RL as the sampling model to evaluate the difficulty of each question in the training sets of GSM8K and MATH, obtaining the fail rate for each question as its difficulty score.
We then use this data to train a difficulty scorer based on DeepSeekMath-7B-Base with mean squared error (MSE) loss.
We then use the scorer to predict the difficulty of each synthetic question.
A portion of overly simple questions is then filtered out, which is empirically determined based on the difficulty distribution of the questions.


\subsection{Response Generation}
\label{subsec:solution_generation}

We use the reward model score as a metric for evaluating the quality of responses.
For each question, we generate 5 solutions and select the solution with the highest reward model scores as the preferred solution.
In our experiments, we use InternLM2-7B-Reward~\citep{cai2024internlm2} as our reward model.
This choice is primarily guided by the model’s performance on the reasoning subset of the Reward Bench~\citep{lambert2024rewardbench}.

\section{Experiment}
\label{sec:experiment}

\begin{table*}[ht]
    \centering
    \small
    \begin{tabular}{l|l|ccccc}
        \toprule
        \bf Model & \bf Synthesis Model & \bf GSM8K & \bf MATH & \bf \makecell{College \\ Math} & \bf \makecell{Olympiad \\ Bench} & \textbf{Average} \\
        \midrule
        \multicolumn{7}{c}{\bf Teacher Models in Data Synthesis} \\
        \midrule
        \gptfourlogo GPT-4-0314 & - & 94.7 & 52.6 & 24.4 & - & - \\
        \gptfourturbologo GPT-4-Turbo-24-04-09 & - & 94.5 & 73.4 & - & - & - \\
        \gptfourologo GPT-4o-2024-08-06 & - & 92.9 & 81.1 & 50.2 & 43.3 & 66.9 \\
        \deepseeklogo DeepSeekMath-7B-RL & - & 88.2 & 52.4 & 41.4 & 19.0 & 49.3 \\
        \qwentwologo Qwen2-Math-7B-Instruct & - & 89.5 & 73.1 & 50.5 & 37.8 & 62.7 \\
        \midrule
        \multicolumn{7}{c}{\bf General Base Model} \\
        \midrule
        Mistral-7B-WizardMath & \gptfourlogowithtext & 81.9 & 33.3 & 21.5 & 8.6 & 36.3 \\
        Mistral-7B-MetaMath & \gptthreepointfivelogowithtext & 77.7 & 28.2 & 19.1 & 5.8 & 32.7 \\
        Mistral-7B-MMIQC & \gptfourlogowithtext & 75.7 & 36.3 & 24.8 & 10.8 & 36.9 \\
        Mistral-7B-MathScale & \gptthreepointfivelogowithtext & 74.8 & 35.2 & 21.8 & - & - \\
        Mistral-7B-KPMath & \gptfourlogowithtext & 82.1 & 46.8 & - & - & - \\
        Mistral-7B-DART-Math & \deepseeklogowithtext & 81.1 & 45.5 & 29.4 & 14.7 & 42.7 \\
        Mistral-7B-NuminaMath & \gptfourologowithtext & 82.1 & 49.4 & 33.8 & 19.4 & 46.2 \\
        \rowcolor{TableRowColor} Mistral-7B-ScaleQuest & \qwentwologowithtext & \textbf{88.5} & \textbf{62.9} & \textbf{43.5} & \textbf{26.8} & \textbf{55.4} \\
        \midrule
        Llama3-8B-MetaMath & \gptthreepointfivelogowithtext & 77.3 & 32.5 & 20.6 & 5.5  & 34.0 \\
        Llama3-8B-MMIQC & \gptfourlogowithtext & 77.6 & 39.5 & 29.5 & 9.6 & 39.1 \\
        Llama3-8B-DART-Math & \deepseeklogowithtext & 81.1 & 46.6 & 28.8 & 14.5 & 42.8 \\
        Llama3-8B-NuminaMath & \gptfourologowithtext & 77.2 & 50.7 & 33.2 & 17.8 & 44.7 \\
        \rowcolor{TableRowColor} Llama3-8B-ScaleQuest & \qwentwologowithtext & \textbf{87.9} & \textbf{64.4} & \textbf{42.8} & \textbf{25.3} & \textbf{55.1} \\
        \midrule
        \multicolumn{7}{c}{\bf Math-Specialized Base Model} \\
        \midrule
        DeepSeekMath-7B-Instruct & - & 82.7 & 46.9 & 37.1 & 14.2 & 45.2 \\
        DeepSeekMath-7B-MMIQC & \gptfourlogowithtext & 79.0 & 45.3 & 35.3 & 13.0 & 43.2 \\
        DeepSeekMath-7B-KPMath-Plus & \gptfourlogowithtext & 83.9 & 48.8 & - & - & - \\
        DeepSeekMath-7B-DART-Math & \deepseeklogowithtext & 86.8 & 53.6 & 40.7 & 21.7 & 50.7 \\
        DeepSeekMath-7B-Numina-Math & \gptfourologowithtext & 75.4 & 55.2 & 36.9 & 19.9 & 46.9 \\
        \rowcolor{TableRowColor} DeepSeekMath-7B-ScaleQuest & \qwentwologowithtext & \textbf{89.5} & \textbf{66.6} & \textbf{47.7} & \textbf{29.9} & \textbf{58.4} \\
        \midrule
        Qwen2-Math-7B-MetaMath & \gptthreepointfivelogowithtext & 83.9 & 49.5 & 39.9 & 17.9 & 47.8 \\
        Qwen2-Math-7B-DART-Math & \deepseeklogowithtext & 88.6 & 58.8 & 45.4 & 23.1 & 54.0 \\
        Qwen2-Math-7B-Numina-Math & \gptfourologowithtext & 84.6 & 65.6 & 45.5 & 33.6 & 57.3 \\
        \rowcolor{TableRowColor} Qwen2-Math-7B-ScaleQuest & \qwentwologowithtext & \textbf{89.7} & \textbf{73.4} & \textbf{50.0} & \textbf{38.5} & \textbf{62.9} \\
        \bottomrule
    \end{tabular}
    \caption{Main results on four mathematical reasoning benchmarks. \textbf{Bold} means the best score with the same base model. The baselines use different synthesis models for both question synthesis and response generation, such as GPT-3.5, GPT-4, and GPT-4o. For our approach, DSMath-7B-QGen and Qwen2-Math-7B-QGen are utilized for question synthesis, with Qwen2-Math-7B-Instruct used for response generation. If multiple models are used, only the latest released one is marked. More details about these datasets are shown in Table~\ref{tab:compared_dataset}.}
    \label{tab:main_experiment}
\end{table*}

\subsection{Experimental Setup}
\label{subsec:experimental_setup}

\paragraph{Training Problem Designers}
In addition to Qwen2-Math-QGen, we also train another question generator based on DeepSeekMath-7B-RL.
We find that combining questions synthesized by multiple generators enhances data diversity, leading to improved final performance, as further discussed in section~\ref{subsec:ablation}.
In this process of the overall data synthesis, many models are mentioned in the context of QFT and QPO.
We explain the inherent insights behind the model selection in Appendix~\ref{sec:insights_behind_model_selection}.


\paragraph{Question \& Response Generation}
The two question generation models are then utilized to generate a total of 2 million questions, with 1 million from each model.
For difficulty filtering, we filter out a portion of overly simple questions generated by \texttt{DeepSeekMath-QGen} and keep the questions generated by \texttt{Qwen2-Math-QGen} as the difficulty distribution is more balanced.
Based on the problems, we synthesize responses (section~\ref{subsec:solution_generation}) using Qwen2-Math-7B-Instruct~\citep{yang2024qwen2}.
For each problem, we sample 5 solutions and select the one with the highest reward score as the final response.
The final dataset consists of 1 million problem-solution pairs.
A detailed analysis of our constructed dataset is provided in Appendix~\ref{sec:appendix_data_statistics}.

\paragraph{Instruction Tuning}
We conduct instruction tuning on the synthetic problems and solutions using two general base models, Mistral-7B~\citep{jiang2023mistral} and Llama3-8B~\citep{dubey2024llama}, as well as two math-specialized base models, DeepSeekMath-7B~\citep{shao2024deepseekmath} and Qwen2-Math-7B~\citep{yang2024qwen2}.
More hyperparameters in the process are provided in Appendix~\ref{sec:hyperparameters}.

\paragraph{Evaluation and Metrics}
We assess the fine-tuned models' performance across four datasets of increasing difficulty.
Along with the widely used GSM8K (elementary level) and MATH (competition level), we include two more challenging benchmarks: College Math~\citep{tang2024mathscale} (college level) and Olympiad Bench~\citep{he2024olympiadbench} (Olympiad level).
For evaluation, we employ the script from \citet{tong2024dart} to extract final answers and determine correctness by comparing answer equivalency.
The generated outputs are all in the form of natural language Chain-of-Thought (CoT) reasoning~\citep{wei2022chain} through greedy decoding, with no tool integration, and we report zero-shot pass@1 accuracy.
We also conduct more comprehensive evaluation, with details shown in Appendix~\ref{sec:more_result}.
We also extend our approach to the long cot and code reasoning task, with details and experiments provided in Appendix~\ref{sec:scalequest_code}.

\paragraph{Compared Baselines}
We mainly compare with data synthesis methods, including question-driven approach such as WizardMath~\citep{luo2023wizardmath}, MetaMath~\citep{yu2023metamath}, MMIQC~\citep{liu2024augmenting}, Orca-Math~\citep{mitra2024orca} and knowledge-driven approach such as KPMath~\citep{huang2024key}, MathScale~\citep{tang2024mathscale}.
In addition to this, we also involve other large math corpus like DART-Math~\citep{tong2024dart} and Numina-Math~\citep{numina_math_datasets}.
More details of these datasets are shown in Table~\ref{tab:compared_dataset}.

\subsection{Main Results}

\paragraph{ScaleQuest significantly outperforms other baselines}
Table~\ref{tab:main_experiment} presents the results. ScaleQuest significantly outperforms previous synthetic methods, with average performance improvements ranging from 5.6\% to 11.5\% over the prior state-of-the-art (SoTA) on both general base models and math-specialized foundation models.
Qwen2-Math-7B-ScaleQuest achieve a zero-shot pass@1 accuracy of 73.4 on the MATH benchmark, matching the performance of GPT-4-Turbo.
For out-of-domain tasks, Qwen2-Math-7B-ScaleQuest outperform its teacher model, Qwen2-Math-7B-Instruct, with scores of 89.7 on the GSM8K benchmark, 73.4 on the MATH benchmark, and 38.5 on the Olympiad benchmark.
In this experiment, we do not strictly control for identical training data volumes due to practical constraints (i.e., some datasets are not publicly available).
To address this, we compare the results with publicly available datasets under equal data volumes in Appendix~\ref{subsec:comparison_under_equal_training_data}.
It’s important to highlight that Qwen2-Math-7B-Instruct has undergone preference alignment, utilizing the powerful reward model Qwen2-Math-RM-72B~\citep{yang2024qwen2}, while our model is only an instruction tuning version.


\paragraph{ScaleQuest scales well with increasing data}

We also compare the scalability of our dataset with other publicly available datasets, including MetaMath~\citep{yu2023metamath}, DART-Math~\citep{tong2024dart}, and Numina-Math~\citep{numina_math_datasets}.
We train Llama3-8B on these datasets and compare their performance changes with increased data size.
The results are presented in Figure~\ref{fig:abstract_results}.
For the in-domain evaluation (MATH), our method demonstrates high data efficiency, achieving superior results with the same amount of data.
In out-of-domain evaluations (Olympiad Bench), it also shows strong scalability, continuing to improve even as other datasets reach their limits.
A limited question set leads to constrained improvements in model performance, as demonstrated by the results of DART-Math, which relies on a small number of questions and generates numerous correct answers through rejection sampling.
Our results further demonstrate that diverse questions support sustained performance growth, emphasizing the need for broader and more varied question generation.

\subsection{Ablation study}
\label{subsec:ablation}

\begin{figure*}[t]
    \centering
    \includegraphics[width=1.0\linewidth]{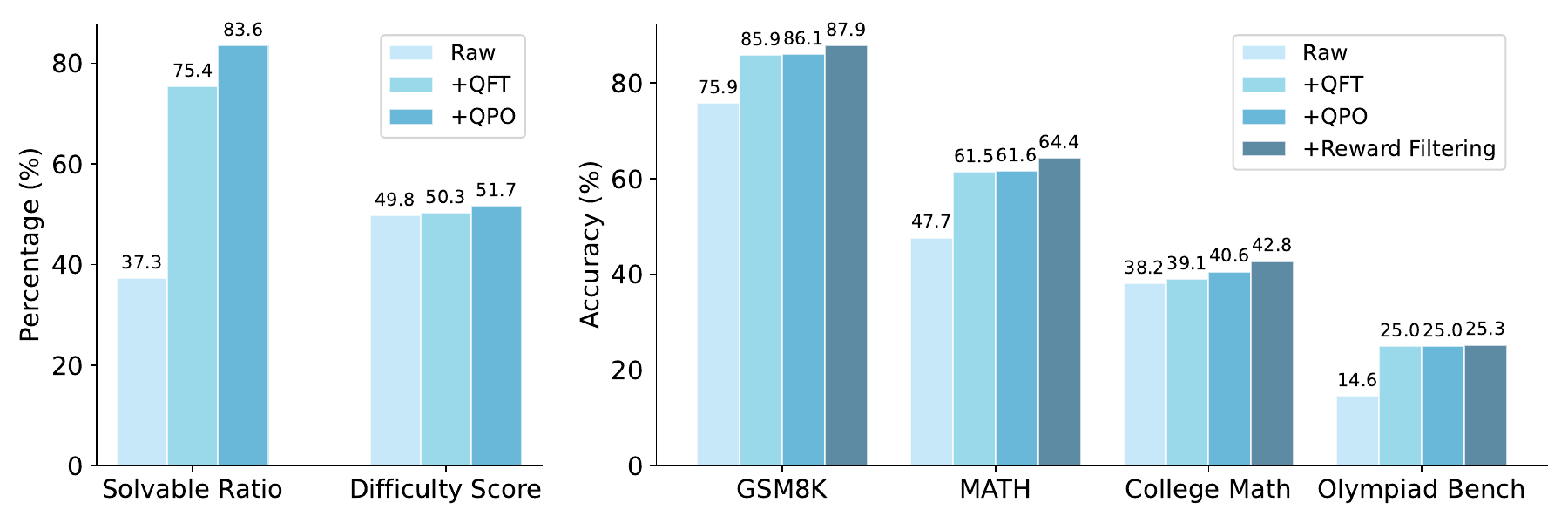}
    \caption{A comparison of the synthetic dataset generated by the raw instruct model, the model after QFT, the model after QPO, and the final dataset after applying reward filtering. \textbf{Left:} The solvable ratio and difficulty score of the generated questions. The solvable ratio refers to the proportion of generated questions that are judged as ``solvable'', while the difficulty score represents the average difficulty rating assigned to each generated question. For difficulty evaluation, we calculate the dataset’s average difficulty score based on ratings for each question: ``very easy'' is rated as 20 points, ``easy'' as 40 points, ``medium'' as 60 points, ``hard'' as 80 points, and ``very hard'' as 100 points. \textbf{Right}: The instruction tuning effectiveness on Llama3-8B-Base.}
    \label{fig:ablation-submethod}
\end{figure*}

\paragraph{Each sub-method contributes to the final performance}
To validate the effectiveness of each of our sub-methods, including QFT, QPO, and reward filtering, we conduct an ablation study.
We evaluate the quality of the questions generated by the models across three dimensions: solvability, difficulty, and performance in instruction tuning.
For solvability and difficulty, we use the same evaluation process as stated in section~\ref{subsec:qpo}.

The results are shown in Figure~\ref{fig:ablation-submethod}.
The ``raw model'' refers to using the instruct model to directly generate instructions and responses, as done in \citet{xu2024magpie}.
To ensure fairness, we also generate 1M question-response pairs using their method based on Qwen2-Math-7B-Instruct, which are used to train Llama3-8B-Base.
After applying QFT and QPO, the model’s performance improves across all three evaluation dimensions, demonstrating the effectiveness of each sub-method.

\begin{table}[t]
    \centering
    \small
    \resizebox{\linewidth}{!}{
    \begin{tabular}{lccccc}
        \toprule
        \makecell{Question \\ Source} & GSM8K & MATH & \makecell{CM} & \makecell{OB} & \textbf{Avg} \\
        \midrule
        MetaMath & 84.5 & 53.8 & 40.1 & 22.1 & 50.1 \\
        OrcaMath & 84.2 & 53.7 & 40.5 & 23.7 & 50.5 \\
        NuminaMath & 86.0 & 65.9 & 46.1 & \textbf{30.2} & 57.1 \\
        ScaleQuest & \textbf{89.5} & \textbf{66.6} & \textbf{47.7} & 29.9 & \textbf{58.4} \\
        \bottomrule
    \end{tabular}}
    \caption{Comparison of question quality across different datasets. To ensure consistency, all responses are generated using Qwen2-Math-7B-Instruct with the same reward filtering process. CM and OB refer to College Math and Olympiad Bench, respectively. Additional results based on the other three backbone models are shown in Appendix~\ref{subsec:different_base_models}.}
    \label{tab:ablation_question}
\end{table}

\paragraph{Comparison of question quality}
To directly compare the question quality of our constructed data with other open-source datasets, we use the same model, Qwen2-Math-7B-Instruct, to generate responses.
We then fine-tune DeepSeekMath-7B-Base using these synthetic datasets.
As shown in Table~\ref{tab:ablation_question}, our model outperforms other synthetic datasets like MetaMath and OrcaMath, highlighting the high quality of our questions.
NuminaMath also demonstrates competitive performance, largely due to the fact that many of its questions are drawn from real-world scenarios.
This highlights the importance of question quality for synthetic data.

\begin{table}[t]
    \centering
    \small
    \resizebox{\linewidth}{!}{
    \begin{tabular}{lccccc}
        \toprule
        Dataset & GSM8K & MATH & CM & OB & \textbf{Avg} \\
        \midrule
        \deepseeklogo SQ-DSMath & 87.6 & 52.2 & 39.8 & 19.4 & 49.8 \\
        \qwentwologo SQ-Qwen2 & 86.8 & 56.1 & 39.6 & 18.7 & 50.3 \\
        Mixed & \textbf{87.8} & \textbf{58.0} & \textbf{40.1} & \textbf{22.2} & \textbf{52.0} \\
        \bottomrule
    \end{tabular}}
    \caption{The performance of Mistral-7B fine-tuned on ScaleQuest-DSMath, ScaleQuest-Qwen2, and a mix of both. We fix the training data size at 400K and find that the mixed data results in the largest improvement.}
    \label{tab:ablation_multimodels}
\end{table}

\begin{table*}[t]
    \small
    \centering
    \begin{tabular}{llcccc}
    \toprule
    \multicolumn{2}{c}{Phase} & Type & \# Samples & GPU hours & Cost (\$) \\
    \midrule
    \multirow{2}{*}{QFT} & Training DSMath-QFT & Train & 15K & 2.0 & 2.6 \\
     & Training Qwen2-Math-QFT & Train & 15K & 1.9 & 2.5 \\
    \midrule
    \multirow{3}{*}{QPO} & Generate Questions & Infer & 10K$\times$2 & 0.4 & 0.5 \\
     & Construct Preference Data & API & 10K$\times$2 & - & 6.2 \\ 
     & QPO Training & Train & 10K$\times$2 & 6.6 & 8.5 \\
     \midrule
     \multirow{4}{*}{Data Synthesis} & Question Generation & Infer & 2M & 38.4 & 49.5 \\
      & solvability \& difficulty check & Infer & 2M & 110.6 & 142.7 \\
      & Response Generation & Infer & 1M$\times$5 & 251.0 & 323.8 \\
      & Reward Scoring & Infer & 1M$\times$5 & 112.0 & 144.5 \\
    \midrule
    \textbf{Total} & & & 1M & 522.9 & 680.8 \\
    \midrule
    \multicolumn{3}{l}{GPT-4 cost (generating the same number of tokens)} & - & - & 24,939.5 \\
    \multicolumn{3}{l}{GPT-4o cost (generating the same number of tokens)} & - & - & 6,115.9 \\
    \bottomrule
    \end{tabular}
    \caption{Cost analysis of the entire data synthesis process. We also estimate the cost of generating the same number of tokens using proprietary models GPT-4 and GPT-4o for comparison.}
    \label{tab:cost_analysis}
\end{table*}

\paragraph{Multiple question generators enhance data diversity}
We use two models as question generators: DSMath-QGen and Qwen2-Math-QGen, which are based on DeepSeekMath~\citep{shao2024deepseekmath} and Qwen2-Math~\citep{yang2024qwen2}, respectively.
To explore the impact of using multiple question generators, we compare the effects of using data synthesized by a single generator versus a mix of data from both.
As shown in Table~\ref{tab:ablation_multimodels}, we find that the mixed data outperforms the data generated by either single generator.
A possible explanation for this improvement is the increased data diversity.
In fact, we observe that DSMath-QGen tends to generate simpler, more real-world-oriented questions, while Qwen2-Math-QGen produces more challenging, theory-driven ones.

\subsection{Human Evaluation Results}
\label{subsec:human_evaluation_results}

We conduct a human evaluation of the generated data, focusing on three aspects: clarity, reasonableness, and real-world relevance. For reference, we also include two high-quality, human-curated datasets, GSM8K and MATH. A total of 40 examples are sampled from each dataset and evaluated based on clarity, coherence, and real-world relevance, with scores ranging from 1 to 5. The results are presented in Table~\ref{tab:human_evaluation}.
In terms of clarity and reasonableness, our synthetic data surpasses NuminaMath but still falls short of the high-quality, real-world datasets like the training sets of GSM8K and MATH.
Regarding real-world relevance, GSM8K leans toward practical, real-life scenarios, while MATH focuses more on theoretical mathematical derivations. Our generated data can be seen as a balance between the two.

\begin{table}[t]
    \centering
    \small
    \begin{tabular}{l|ccc}
        \toprule
        \bf Dataset & \bf Clarity & \bf Reasonableness & \bf \makecell{Real-world \\ relevance} \\
        \midrule
        GSM8K & 4.4 & 4.5 & 3.9 \\
        MATH & 4.1 & 4.3 & 2.4 \\
        NuminaMath & 3.8 & 4.0 & 2.4 \\
        \midrule
        ScaleQuest & 3.9 & 4.0 & 2.8 \\
        \bottomrule
    \end{tabular}
    \caption{Human Evaluation Results.}
    \label{tab:human_evaluation}
\end{table}

\subsection{Cost Analysis}
\label{subsec:cost_analysis}

The data synthesis process is conducted on a server with 8 A100-40G-PCIe GPUs.
We summarize our overall costs in Table~\ref{tab:cost_analysis}.
Generating 1 million data samples requires only 522.9 GPU hours (approximately 2.7 days on an 8-GPU server), with an estimated cost of \$680.8 for cloud server rental.\footnote{\url{https://lambdalabs.com/service/gpu-cloud}}
This is only about 10\% of the cost of generating the same data using GPT-4o, demonstrating the cost-effectiveness of our method.

\section{Related Work}
\label{sec:related_work}

\subsection{Mathematical Reasoning}

Solving math problems is regarded as a key measure of evaluating the reasoning ability of LLMs.
Recent advancements in mathematical reasoning for LLMs, including models like OpenAI o1, Claude-3.5, Gemini~\citep{reid2024gemini}, DeepSeekMath~\citep{shao2024deepseekmath}, InternLM2-Math~\citep{cai2024internlm2}, and Qwen2.5-Math~\citep{yang2024qwen25}, have spurred the development of various approaches to improve reasoning capabilities of LLMs on math-related tasks.
To strengthen the math reasoning capabilities of LLMs, researchers have focused on areas such as prompting techniques~\citep{chia2023contrastive,chen2023skills,zhang2023cumulative}, data construction for pretraining~\citep{lewkowycz2022solving,azerbayev2023llemma,zhou2024jiuzhang3,shao2024deepseekmath} and instruction tuning~\citep{luo2023wizardmath,yue2023mammoth}, tool-integrated reasoning\citep{chen2022program,gao2023pal,gou2023tora,wang2023mathcoder,yue2024mammoth2,yin2024mumath,zhang2024evaluating}, and preference tuning~\citep{ma2023let,luong2024reft,shao2024deepseekmath,lai2024step}.
Our work primarily focuses on math data synthesis for instruction tuning.

\subsection{Data Synthesis for Instruction Tuning}
High-quality reasoning data, particularly well-crafted questions, is in short supply.
Prior efforts have mostly started with a small set of human-annotated seed instructions and expanded them through few-shot prompting.
We categorize them into two types: question-driven augmentation~\citep{luo2023wizardmath,yu2023metamath,li2024common,liu2024augmenting,li2024synthetic} and knowledge-driven augmentation~\citep{didolkar2024metacognitive,shah2024ai,tang2024mathscale,huang2024key}.
There are other methods for enhancing dataset quality as well.
DART-Math~\citep{tong2024dart} focuses on enhancing the quality of responses by using difficulty-aware rejection sampling.
In contrast, Numina-Math~\citep{numina_math_datasets} improves its dataset by collecting more real-world and synthetic data.
These high-quality datasets can be integrated with our constructed dataset, resulting in an improved data mix for more effective instruction tuning.

\section{Conclusion}
\label{sec:conclusion}

In this work, we propose ScaleQuest, a novel data synthesis framework that unlocks the ability of lightweight models to independently generate large-scale, high-quality reasoning data from scratch, at a low cost. 
Using this dataset, we fine-tune the model and achieve remarkable improvements, with gains ranging from 29.2\% to 46.4\% compared to the base model, outperforming the strong baselines.

\clearpage
\section*{Limitation}
We explore the potential of lightweight open-sourced models to generate high-quality instruction tuning data.
However, the effectiveness on larger and more powerful models, such as Qwen2.5-Math-72B-Instruct and Llama3.3-70B-Instruct, remains uncertain.
Our future research will focus on experimenting with these larger models.
Additionally, although the optimizations for solvability and difficulty have shown positive effects, the model-generated responses still fall short of being fully satisfactory.
Therefore, further improvements in question preference alignment are crucial and will be an important direction for our future work.

\section*{Acknowledgements}
We want to thank all the anonymous reviewers for their valuable comments. This work was supported by the National Science Foundation of China (NSFC No. 62206194), the Natural Science Foundation of Jiangsu Province, China (Grant No. BK20220488), the Young Elite Scientists Sponsorship Program by CAST (2023QNRC001), the Priority Academic Program Development of Jiangsu Higher Education Institutions, and Sponsored by CCF-Tencent Rhino-Bird Open Research Fund. We also acknowledge MetaStone Tech. Co. for providing us with the software, optimisation on high performance computing and computational resources required by this work.


\bibliography{custom}

\clearpage
\appendix


\section{Hyper-parameters settings}
\label{sec:hyperparameters}

\paragraph{Training Problem Designers}

During the QFT stage, both models are trained on a mixed training subset of GSM8K and MATH problems, containing a total of 15K problems.
We train for only 1 epoch, considering that training for more epochs might cause the models to overfit the training problems and negatively impact the diversity of generated questions.
We also use sequence packing~\citep{krell2021efficient} to accelerate training.
In the QPO stage, we use 10K preference data for training, with a learning rate of 5e-7 and a batch size of 128.

\paragraph{Question Generation}

During this process, we set the maximum generation length to 512, a temperature of 1.0, and a top-p value of 0.99.
To ensure quality, we apply a question filtering pipeline (section~\ref{subsec:question_filtering}) that involves language filtering, solvability filtering, and difficulty sampling.
This process refines the dataset, leaving approximately 1M questions, 400K from \texttt{DeepSeek-QGen} and 600K from \texttt{Qwen2-Math-QGen}.

\paragraph{Response Generation}

In the process, we set the maximum generation length to 2048, with a temperature of 0.7 and top-p of 0.95.
We use chain-of-thought prompt~\citep{wei2022chain} to synthesize solutions.
We use vLLM~\citep{kwon2023efficient} to accelerate the generation and Ray~\citep{moritz2018ray} to deploy distributed inference.

\paragraph{Instruction Tuning}

All models are fine-tuned for 3 epochs in our experiments unless specified otherwise. We use a linear learning rate schedule with a 3\% warm-up ratio, reaching a peak of 5e-5 for Llama3 and DeepSeekMath and 1e-5 for the other models, followed by cosine decay to zero.

\section{Additional Data Statistics}
\label{sec:appendix_data_statistics}

\paragraph{Filtering process}

\begin{figure}[ht]
    \centering
    \includegraphics[width=1.0\linewidth]{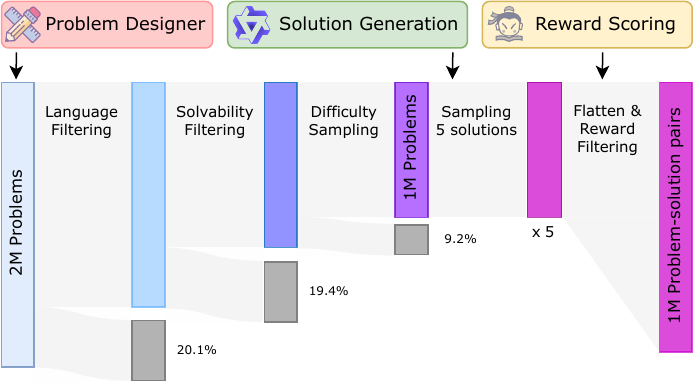}
    \caption{Overview of our filtering process.}
    \label{fig:data_processing}
\end{figure}

The entire data generation process is illustrated in Figure~\ref{fig:data_processing}.
After using the two question generators to produce 2 million questions from scratch, we perform a filtering process, including language filtering, solvability checks, and difficulty sampling.
These steps filter out 20.1\%, 19.4\%, and 9.2\% of the samples, respectively, resulting in a final question set of 1 million questions.
In the subsequent response generation process, we filter out responses without answers by checking for key phrases such as ``The answer is'' or ``\textbackslash boxed\{\}''.
This step eliminates a negligible portion of the samples, as most of the filter questions are solvable and did not pose any confusion for the response generation model.

\paragraph{Dataset Coverage}

\begin{figure*}[ht]
    \centering
    \begin{minipage}[b]{0.35\textwidth}
        \centering
        \includegraphics[width=\linewidth]{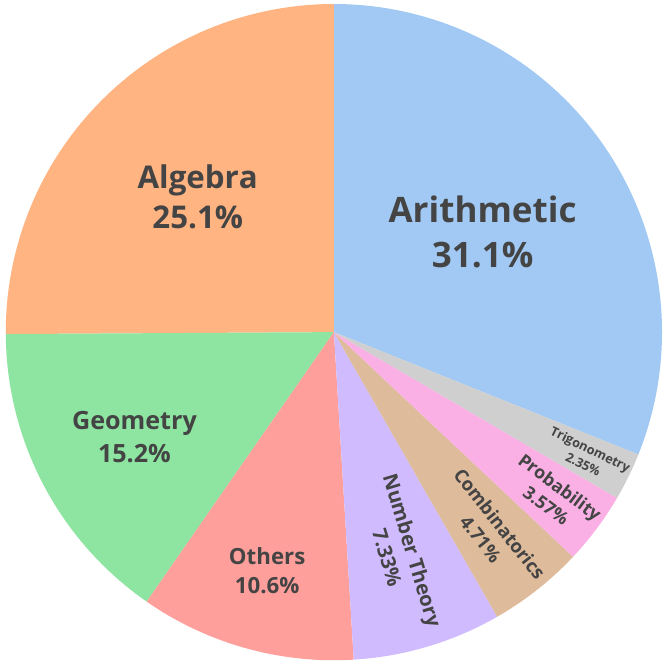}
        \caption{Topic distribution of our generated dataset.}
        \label{fig:topic_coverage}
    \end{minipage}
    \hspace{0.05\textwidth} 
    \begin{minipage}[b]{0.55\textwidth}
        \centering
        \includegraphics[width=\linewidth]{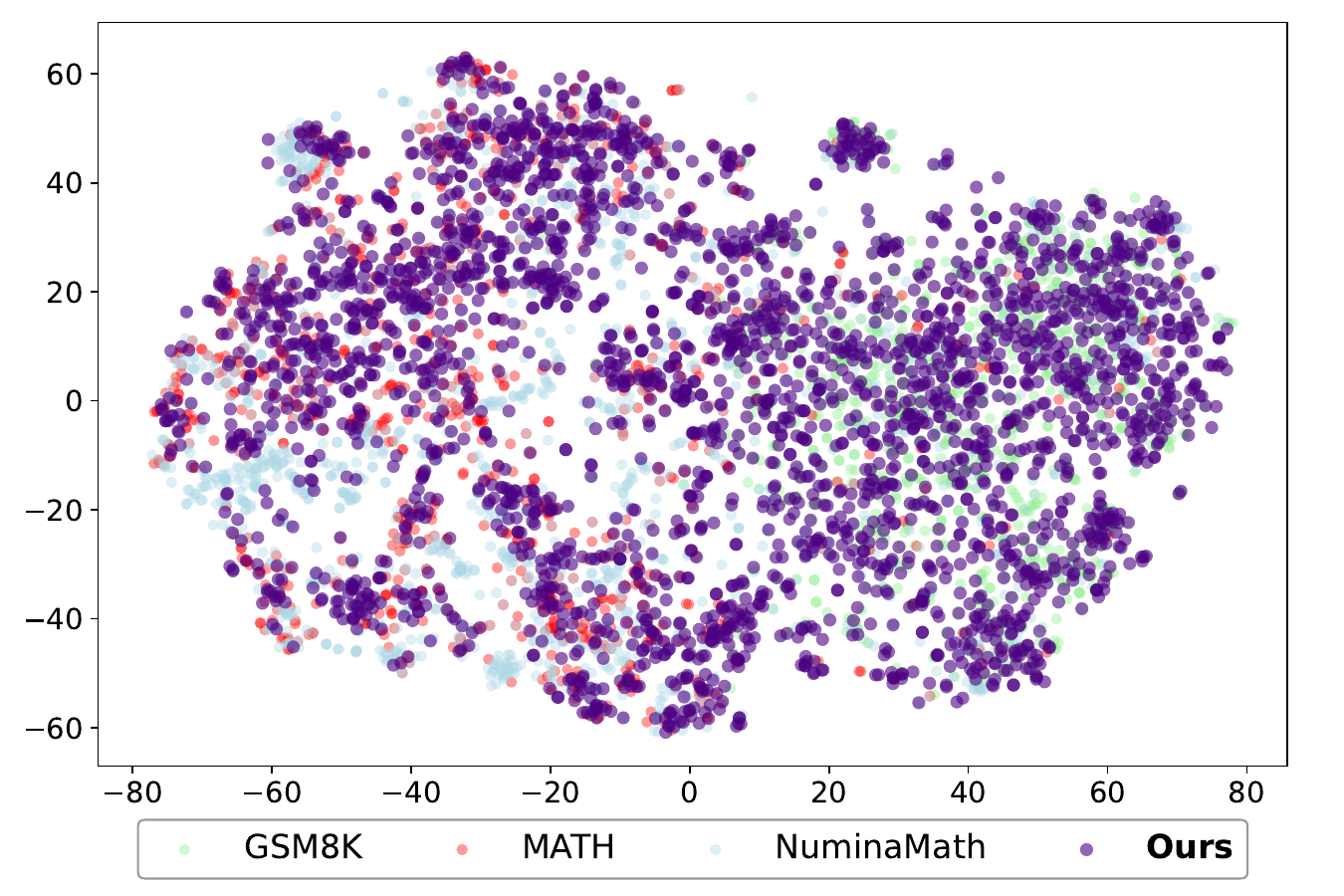}
        \caption{t-SNE plot of our dataset, with GSM8K, MATH, and NuminaMath.}
        \label{fig:embedding_coverage}
    \end{minipage}
\end{figure*}

We analyze the dataset coverage through two aspects:
(1) Problem Topic Coverage, such as algebra and geometry.
Following \citet{huang2024key}, we use GPT-4o to categorize the topics of the given questions, with prompt illustrated in Figure~\ref{fig:prompt_topic}.
Figure~\ref{fig:topic_coverage} presents the results.
We find that the topics cover the major areas of mathematics, such as arithmetic, algebra, geometry, and others.
(2) Embedding space analysis. Following \citet{zhao2024wildchat} and \citet{xu2024magpie}, we first compute the input embeddings of the questions and then project them into a two-dimensional space using t-SNE~\citep{van2008visualizing}.
We include only real-world datasets, such as GSM8K~\citep{cobbe2021training}, MATH~\citep{hendrycks2021measuring}, and NuminaMath~\citep{numina_math_datasets} (which contains a small portion of synthetic questions). As shown in Figure~\ref{fig:embedding_coverage}, our synthetic data closely resembles the real-world questions.

\paragraph{Data Leakage Analysis}
We conduct an n-gram similarity analysis between the generated questions and all test sets from both our dataset and other baseline datasets.
Based on prior empirical analysis~\citep{brown2020language,wei2021finetuned}, we set n=13 to prevent spurious collisions and calculate how much the test sets overlap with training data to assess data contamination.
Table~\ref{tab:data_leak} shows the clean ratio of our dataset and other baseline datasets.
The results demonstrate that our dataset achieves a relatively high level of data cleanliness compared to other datasets, suggesting that our method generates novel questions instead of memorizing existing ones.

\begin{table}[ht]
    \centering
    \small
    \resizebox{\linewidth}{!}{
    \begin{tabular}{l|ccccc}
        \toprule
        \textbf{Dataset} & GSM8K & MATH & CM & OB & \textbf{Avg} \\
        \midrule
        MetaMath & 99.8 & 92.2 & 100 & 99.7 & 97.9 \\
        NuminaMath & 99.8 & 89.8 & 99.9 & 86.8 & 94.1 \\
        DART-Math & 99.8 & 91.5 & 100.0 & 99.6 & 97.7 \\
        MMIQC & 99.8 & 88.0 & 98.9 & 97.9 & 96.2 \\
        \midrule
        SQ (Ours) & 99.9 & 92.8 & 99.8 & 97.2 & 97.4 \\
        \bottomrule
    \end{tabular}}
    \caption{Overlap statistics for the datasets used. We report the clean ratio of the test set, representing the percentage of test samples that have no matching n-grams with samples in the training set.}
    \label{tab:data_leak}
\end{table}

\paragraph{Safety Analysis}

We use Llama3-8B-Guard~\citep{inan2023llama} as a discriminator model to detect any unsafe elements in the data. After sampling 10K instances from the 1 million samples, we find that only 0.1\% are flagged as unsafe.

\paragraph{Generated Examples}
We sample several generated examples from our datasets, as shown in Figure~\ref{fig:scalequest_example1}, \ref{fig:scalequest_example2} and \ref{fig:scalequest_example3}.
The generated math problems are of high quality, driving effective learning.

\begin{table*}[ht]
    \centering
    \small
    \begin{tabular}{lrlc}
        \toprule
        \textbf{Dataset} & Size & Synthesis Model & \makecell{Public} \\
        \midrule
        WizardMath~\citep{luo2023wizardmath} & 96K & GPT-4 & \ding{55} \\ 
        MetaMath~\citep{yu2023metamath} & 395K & GPT-3.5-Turbo & \ding{51} \\
        MMIQC~\citep{liu2024augmenting} & 2294K & GPT-4 \& GPT-3.5-Turbo \& Human & \ding{51} \\
        Orca-Math~\citep{mitra2024orca} & 200K & GPT-4-Turbo & \ding{51} \\
        Xwin-Math~\citep{li2024common} & 1440K & GPT-4-Turbo & \ding{55} \\
        KPMath-Plus~\citep{huang2024key} & 1576K & GPT-4 & \ding{55} \\
        MathsScale~\citep{tang2024mathscale} & 2021K & GPT-3.5 \& Human & \ding{55} \\
        DART-Math~\citep{tong2024dart} & 585K & DeepSeekMath-7B-RL & \ding{51} \\
        Numina-Math~\citep{numina_math_datasets} & 860K & GPT-4 \& GPT-4o & \ding{51} \\
        \midrule
        \multirow{2}{*}{ScaleQuest} & \multirow{2}{*}{1000K} & DeepSeekMath-7B-RL & \multirow{2}{*}{\ding{51}} \\
        & & Qwen2-Math-7B-Instruct &  \\
        \bottomrule
    \end{tabular}
    \caption{Comparison between our constructed dataset and previous datasets.}
    \label{tab:compared_dataset}
\end{table*}

\section{ScaleQuest for Broader tasks}
\label{sec:scalequest_code}

\subsection{Long Chain-of-Thought Mathematical Reasoning}
DeepSeek-R1~\citep{guo2025deepseek} demonstrated the effectiveness of Long Chain-of-Thought (CoT) reasoning in tackling challenging mathematical reasoning tasks.
In this part, we further investigate the capability of ScaleQuest to generate high-quality long CoT data.

\paragraph{Settings}
We utilize the latest Qwen2.5-Math-7B model~\citep{yang2024qwen25} for question fine-tuning and preference optimization.
To construct long CoT data, we select DeepSeek-R1-Distill-Qwen-7B as the response generator.
The rest of the pipeline remains unchanged from prior work.
In total, we synthesized 150K Long Chain-of-Thought training samples.

\paragraph{Results}
We then fine-tuned DeepSeek-R1-Distill-Qwen-7B on this dataset, which also serves as the teacher model during data synthesis.
Table~\ref{tab:math_longcot} presents the results, showing that the fine-tuned model outperforms its teacher, demonstrating a strong potential for self-improvement.

\begin{table*}[ht]
    \centering
    \small
    \begin{tabular}{l|cccc|c}
        \toprule
        Model & GSM8k & MATH500 & AIME24 & AMC25 & Average \\
        \midrule
        DS-R1-Distill-Qwen-7B & 91.5 & 90.2 & 43.3 & 75.0 & 75.0 \\
        DS-R1-Distill-Qwen-7B-ScaleQuest & \textbf{93.0} & \textbf{91.0} & \textbf{53.3} & \textbf{90.0} & \textbf{81.8} \\
        \bottomrule
    \end{tabular}
    \caption{Results of ScaleQuest-LongCoT.}
    \label{tab:math_longcot}
\end{table*}

\subsection{Code Reasoning}
We also extend our ScaleQuest method to the Code Reasoning Task as a simple validation.
We made the following modifications to adapt to the code reasoning task:
\paragraph{Settings}
We choose DeepSeek-Coder-7B-Instruct~\citep{guo2024deepseek} and Qwen2.5-Coder-7B-Instruct~\citep{hui2024qwen2coder} as two problem-solving models to perform question fine-tuning on 20K questions randomly sampled from CodeFeedBack~\citep{zheng2024opencodeinterpreter}.
For Question Preference Optimization, we also focus on solvability and difficulty, making slight modifications to the prompts based on the code reasoning task.
Our evaluation cover HumanEval~\citep{chen2021evaluating}, MBPP~\citep{austin2021program}, and BigCodeBench~\citep{zhuo2024bigcodebench}, using the same evaluation script as Qwen2.5-Coder. We report pass@1 results using greedy search.

\begin{table*}[ht]
    \centering
    \small
    \begin{tabular}{l|c|cccc}
        \toprule
        Model & \# Samples (K) & HumanEval & MBPP & BigCodeBench & Average \\
        \midrule
        Qwen2.5-Coder-CFB & 156 & 79.3 & 77.2 & 35.6 & 64.0 \\
        Qwen2.5-Coder-CFB-Aug & 156 & 84.1 & \textbf{84.7} & 39.0 & 69.3 \\
        Qwen2.5-Coder-ScaleQuest & 156 & \textbf{86.6} & 83.1 & \textbf{40.0} & \textbf{69.9} \\
        \bottomrule
    \end{tabular}
    \caption{Results of ScaleQuest in Code Reasoning Task. All results are based on Qwen2.5-Coder-7B-Base. CFB refers to the CodeFeedBack-Filtered Dataset. We augment the responses for the problems in CodeFeedback-Filtered using Qwen2.5-Coder-7B-Instruct with reward filtering, creating a new dataset referred to as CFB-Aug.}
    \label{tab:code_reasoning}
\end{table*}

\paragraph{Results}
The results are presented in Table~\ref{tab:code_reasoning}.
Compared to the widely used refined version of CodeFeedback, namely CodeFeedback-Filtered, our generated data outperforms it, with an average improvement of 5.9 across the three baselines.
Additionally, we enhance the Response portion of CodeFeedback-Filtered using Qwen2.5-Coder-7B-Instruct, and the results indicate that our generated questions are of higher quality. This further demonstrates the effectiveness of the ScaleQuest method.

\section{Insights behind model selection}
\label{sec:insights_behind_model_selection}

In our works, we use multiple models, e.g., DSMath-7B-RL, Qwen2-Math-7B-Ins, GPT-4o-mini, and DSMath-7B-Base, which may cause confusion for model selection.
In response, we also supplement our approach with a simpler setup.
We use Qwen2-Math-7B-Ins for training question generators, constructing optimization data for QPO, and performing solvability \& difficulty filtering, as well as for response generation. For reward filtering, InternLM-7B-Reward remained unchanged.
The results, as shown in Figure~\ref{tab:same_volumn_results} (ScaleQuest-Simple result), indicate that our approach continues to demonstrate superior performance compared to existing datasets.
Additionally, we summarize these insights on model selection for domain adaptation:
\begin{itemize}[leftmargin=*]
    \setlength{\itemsep}{0pt}
    \setlength{\parskip}{0pt}
    \item \textbf{Selection of base model for training question generator:} The self-synthesis generation paradigm heavily relies on the inherent knowledge of the problem-solving model itself~\citep{xu2024magpie}. Therefore, a domain-specific model is essential. For example, Qwen2-Math-Ins is suitable for mathematical reasoning, while Qwen2.5-Coder-Ins fits well for code reasoning. Furthermore, using multiple question generators often leads to more diverse and higher-quality questions (as discussed in section~\ref{subsec:ablation}).
    \item \textbf{Selection of model for constructing optimization data:} Well-aligned, general-purpose models, such as Llama3.1-70B and GPT-4o-mini, tend to perform better than domain-specific models, as illustrated in Figure~\ref{fig:solvability_and_difficulty}.
    \item \textbf{Selection of Response Generation Model \& Reward Model:} These can be selected based on their performance on the corresponding mathematical tasks.
\end{itemize}
We believe that the methodology and the experience in selecting models are always more critical than the chosen models themselves. With the continuous advancements in the open-source community, we are confident that stronger models will undoubtedly produce even better datasets when applying our approach.

\section{Additional Experiments}
\label{sec:more_result}

\subsection{Evaluation Results on More Out-of-Domain (OOD) Benchmarks}
\label{subsec:results_on_other_ood_benchmarks}
In addition to College Math and Olympiad Bench, we include two additional OOD benchmarks: GSM-Hard~\citep{gao2023pal} and MathChat~\citep{liang2024mathchat}.
GSM-Hard is constructed by modifying the questions in GSM8K, replacing the numbers with larger, less common ones.
From MathChat, we select two problem-solving tasks: follow-up QA and error correction.
The results are summarized in Table~\ref{tab:other_ood_results}.
In more fine-grained OOD evaluations, our model continues to perform on par with Qwen2-Math-7B-Ins, further demonstrating our ScaleQuest Model's generalization capability and highlighting the generated data's robustness.

\begin{table*}[ht]
    \centering
    \small
    \begin{tabular}{l|ccccc|c}
        \toprule
        \multirow{2}{*}{Model} & \multirow{2}{*}{GSM-Hard} & \multicolumn{3}{c}{Follow-up QA} & \multirow{2}{*}{Error Correction} & \multirow{2}{*}{Average} \\
        & & R1 & R2 & R3 & & \\
        \midrule
        Qwen2-Math-7B-Instruct & 68.3 & 89.5 & 62.4 & 53.5 & 89.9 & 72.7 \\
        Qwen2-Math-7B-ScaleQuest & 66.3 & 89.7 & 61.7 & 53.5 & 91.1 & 72.5 \\
        \bottomrule
    \end{tabular}
    \caption{The comparison between Qwen2-Math-7B-Ins and the ScaleQuest Model on GSM-Hard and MathChat. We choose Follow-up QA and Error Correction from MathChat for evaluation in problem-solving. R1, R2, and R3 represent different rounds in Follow-up QA.}
    \label{tab:other_ood_results}
\end{table*}

\subsection{Comparison Under Equal Training Data Volume}
\label{subsec:comparison_under_equal_training_data}

In the right panel of Figure~\ref{fig:abstract_results}, we plot the scaling trends of model performance with increasing data volume, showcasing the superiority of the ScaleQuest method when using the same amount of data.
To further ensure a fair comparison, we randomly sample the same number of training examples from open-source datasets for training.
Specifically, we sample 400K examples from MetaMath, DART-Math, NuminaMath, and our dataset (for MetaMath, which contains 395K examples in total, all samples are used).
The results are presented in Table~\ref{tab:same_volumn_results}.
We observe that with the same amount of training data, our dataset demonstrates significantly higher instruction tuning effectiveness compared to other datasets.

\begin{table*}[ht]
    \centering
    \small
    \begin{tabular}{l|c|ccccc}
        \toprule
        \textbf{Model} & \# Samples (K) & GSM8K & MATH & \makecell{College \\ Math} & \makecell{Olympiad \\ Bench} & Average \\
        \midrule
        Qwen2-Math-7B-MetaMath & 395 & 84.3 & 48.6 & 40.5 & 15.6 & 47.3 \\
        Qwen2-Math-7B-DART-Math & 400 & 88.6 & 58.2 & 45.2 & 22.8 & 53.7 \\
        Qwen2-Math-7B-NuminaMath & 400 & 82.0 & 65.8 & 44.9 & 29.2 & 55.5 \\
        \midrule
        Qwen2-Math-7B-ScaleQuest & 400 & \textbf{90.6} & \textbf{71.6} & \textbf{50.2} & \textbf{36.2} & \textbf{62.1} \\
        Qwen2-Math-7B-ScaleQuest-Simple & 400 & 89.4 & 69.9 & 48.8 & 33.6 & 60.4 \\
        \bottomrule
    \end{tabular}
    \caption{Results on four mathematical reasoning benchmarks. All results are based on Qwen2-Math-7B-Base. ScaleQuest-Simple is a simplified version that only utilizes Qwen2-Math-7B-Ins for QFT, QPO, and question filtering, and InternLM-7B-Reward for reward filtering.}
    \label{tab:same_volumn_results}
\end{table*}

\begin{table*}[ht]
    \centering
    \small
    \begin{tabular}{l|ccccc}
        \toprule
        Model & GSM8K & MATH & \makecell{College \\ Math} & \makecell{Olympiad \\ Bench} & \textbf{Average} \\
        \midrule
        Mistral-7B-MetaMath-Aug & 77.0 & 34.1 & 18.6 & 8.6 & 34.6 \\
        Mistral-7B-OrcaMath-Aug & 84.4 & 31.6 & 20.9 & 8.2 & 36.3 \\
        Mistral-7B-NumiMath-Aug & 79.5 & 62.8 & 40.4 & \textbf{30.4} & 53.3 \\
        Mistral-7B-ScaleQuest & \textbf{88.5} & \textbf{62.9} & \textbf{43.5} & 28.8 & \textbf{55.9} \\
        \midrule
        Llama3-8B-MetaMath-Aug & 77.6 & 33.1 & 20.6 & 9.2 & 35.1 \\
        Llama3-8B-OrcaMath-Aug & 83.2 & 32.6 & 19.4 & 8.6 & 36.0 \\
        Llama3-8B-NumiMath-Aug & 79.1 & 62.9 & 39.3 & \textbf{25.4} & 51.7 \\
        Llama3-8B-ScaleQuest & \textbf{87.9} & \textbf{64.4} & \textbf{42.8} & 25.3 & \textbf{55.1} \\
        \midrule
        Qwen2-Math-7B-MetaMath-Aug & 88.5 & 68.5 & 47.1 & 33.0 & 59.3 \\
        Qwen2-Math-7B-OrcaMath-Aug & 89.3 & 68.3 & 46.6 & 31.9 & 59.0 \\
        Qwen2-Math-7B-NumiMath-Aug & 89.5 & 72.6 & 49.5 & 36.3 & 62.0 \\
        Qwen2-Math-7B-ScaleQuest & \textbf{89.7} & \textbf{73.4} & \textbf{50.0} & \textbf{38.5} & \textbf{62.9} \\
        \bottomrule
    \end{tabular}
    \caption{Additional results of Table~\ref{tab:ablation_question} on the other base models. All responses are generated using Qwen2-Math-7B-Instruct with the same reward filtering process. For baseline datasets, ``-Aug'' indicates that the responses have been enhanced.}
    \label{tab:ablation_question_more}
\end{table*}

\subsection{Training Data Volume on QPO}
\label{subsec:training_data_volumn_qpo}

\begin{figure}[ht]
    \centering
    \includegraphics[width=1.0\linewidth]{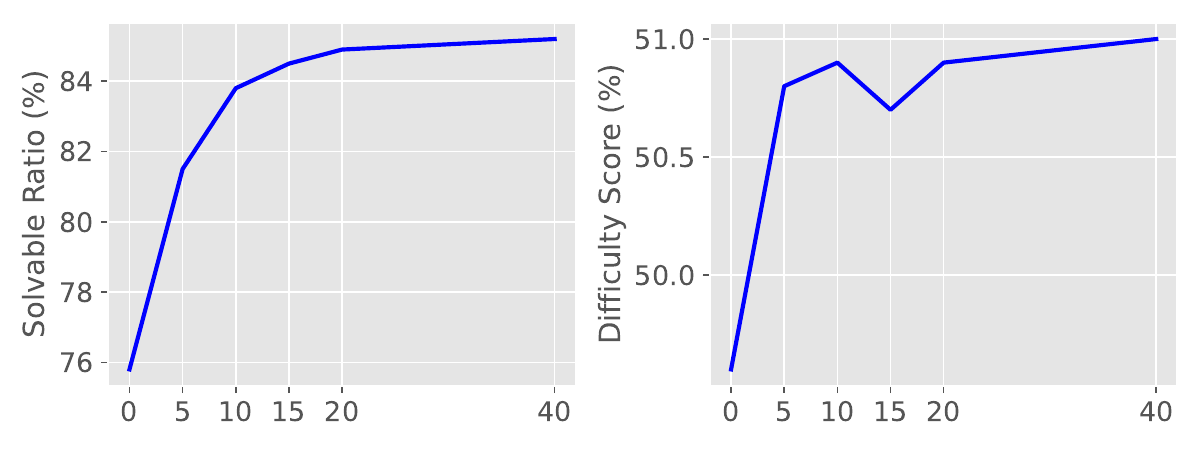}
    \caption{Performance of QPO in different training data volume. The evaluation covers the solvable ratio and difficulty score, following the same evaluation procedure as in Figure~\ref{fig:ablation-submethod}.}
    \label{fig:qpo_data_scale}
\end{figure}

QPO is designed to enhance the solvability and difficulty of the question generator.
We investigate the impact of training data volume by using GPT-4o-mini as the optimization model. The training data volume is controlled at 5K, 10K, 15K, 20K, and 40K, with Qwen2-Math-7B-QFT serving as the base model.
We evaluate the performance of the trained question generator in terms of solvability and difficulty.
The results are shown in Figure~\ref{fig:qpo_data_scale}.
As the amount of training data increases, both the solvable rate and difficulty of the questions generated by the question generator improve, gradually converging around 20K training examples.
We believe that maintaining the training data at approximately 10K represents a more suitable balance between training cost and model performance.

\subsection{Additional Results Based on Different Base Models}
\label{subsec:different_base_models}

We have supplemented Table~\ref{tab:ablation_question} with the results for the other three base models, as shown in Table~\ref{tab:ablation_question_more}.
Under the same response generation process, our approach consistently outperforms existing datasets across all four base models, further demonstrating the superiority of our method.

\section{Prompt Template}
\label{sec:appexdix_prompts}

\begin{figure*}[ht]
\begin{tcolorbox}[
  title=Prompts for Problem Solvability Optimization,
  colback=white,
  colframe=black!70,
  fontupper=\ttfamily,
  coltitle=white,
  colbacktitle=blue!50!black,
  fonttitle=\ttfamily,
  boxrule=0.8pt,
]
Please act as a professional math teacher. \\
Your goal is to create high quality math word problems to help students learn math. \\
You will be given a math question. Please optimize the Given Question and follow the instructions. \\
To achieve the goal, please follow the steps: \\
\# Please check that the given question is a math question and write detailed solution to the Given Question. \\
\# Based on the problem-solving process, double check the question is solvable. \\
\# If you feel that the given question is not a meaningful math question, rewrite one that makes sense to you. Otherwise, modify the Given question according to your checking comment to ensure it is solvable and of high quality. \\
\# If the question can be solved with just a few simple thinking processes, you can rewrite it to explicitly request multiple-step reasoning. \\
\\
You have five principles to do this: \\
\# Ensure the optimized question only asks for one thing, be reasonable and solvable, be based on the Given Question (if possible), and can be answered with only a number (float or integer). For example, DO NOT ask, `what is the amount of A, B and C?'. \\
\# Ensure the optimized question is in line with common sense of life. For example, the amount someone has or pays must be a positive number, and the number of people must be an integer. \\
\# Ensure your student can answer the optimized question without the given question. If you want to use some numbers, conditions or background in the given question, please restate them to ensure no information is omitted in your optimized question. \\
\# Please DO NOT include solution in your question. \\

Given Question: \fcolorbox{TextBackColor}{TextBackColor}{problem} \\
Your output should be in the following format: \\
CREATED QUESTION: [your created question] \\
VERIFICATION AND MODIFICATION: [solve the question step-by-step and modify it to follow all principles] \\
FINAL QUESTION: [your final created question]
\end{tcolorbox}
\caption{Prompts used to optimize the solvability of questions for QPO Training.}
\label{fig:prompt_solvability}
\end{figure*}

\begin{figure*}
\begin{tcolorbox}[
  title=Prompts for Problem Difficulty Optimization,
  colback=white,
  colframe=black!70,
  fontupper=\ttfamily,
  coltitle=white,
  colbacktitle=blue!50!black,
  fonttitle=\ttfamily,
  boxrule=0.8pt,
]
You are an Math Problem Rewriter that rewrites the given \#Problem\# into a more complex version. \\
Please follow the steps below to rewrite the given "\#Problem\#" into a more complex version. \\
\\
Step 1: Please read the "\#Problem\#" carefully and list all the possible methods to make this problem more complex (to make it a bit harder for well-known AI assistants such as ChatGPT and GPT4 to handle). Note that the problem itself might be erroneous, and you need to first correct the errors within it. \\
Step 2: Please create a comprehensive plan based on the \#Methods List\# generated in Step 1 to make the \#Problem\# more complex. The plan should include several methods from the \#Methods List\#. \\
Step 3: Please execute the plan step by step and provide the \#Rewritten Problem\#. \#Rewritten Problem\# can only add 10 to 20 words into the "\#Problem\#". \\
Step 4: Please carefully review the \#Rewritten Problem\# and identify any unreasonable parts. Ensure that the \#Rewritten Problem\# is only a more complex version of the \#Problem\#. Just provide the \#Finally Rewritten Problem\# without any explanation and step-by-step reasoning guidance. \\
\\
Please reply strictly in the following format: \\
Step 1 \#Methods List\#: \\
Step 2 \#Plan\#: \\
Step 3 \#Rewritten Problem\#: \\
Step 4 \#Finally Rewritten Problem\#: \\
\\
\#Problem\#: \fcolorbox{TextBackColor}{TextBackColor}{Problem}
\end{tcolorbox}
\caption{Prompts used to optimize the difficulty of questions for QPO Training.}
\label{fig:prompt_difficulty}
\end{figure*}

\begin{figure*}
\begin{tcolorbox}[
  title=Prompts for Problem Solvability Check,
  colback=white,
  colframe=black!70,
  fontupper=\ttfamily,
  coltitle=white,
  colbacktitle=blue!50!black,
  fonttitle=\ttfamily,
  boxrule=0.8pt,
]
Please act as a professional math teacher. \\
Your goal is to determine if the given problem is a valuable math problem. You need to consider two aspects: \\
1.	The given problem is a math problem. \\
2.	The given math problem can be solved based on the conditions provided in the problem (You can first try to solve it and then judge its solvability). \\
\\
Please reason step by step and conclude with either `Yes' or `No'. \\
\\
Given Problem: \fcolorbox{TextBackColor}{TextBackColor}{Problem}
\end{tcolorbox}
\caption{Prompts used to check the solvability of questions.}
\label{fig:prompt_solvability_check}
\end{figure*}

\begin{figure*}
\begin{tcolorbox}[
  title=Prompts for Difficulty Classification,
  colback=white,
  colframe=black!70,
  fontupper=\ttfamily,
  coltitle=white,
  colbacktitle=blue!50!black,
  fonttitle=\ttfamily,
  boxrule=0.8pt,
]
\# Instruction \\
\\
You first need to identify the given user intent and then label the difficulty level of the user query based on the content of the user query. \\
\\
\#\# User Query \\
\fcolorbox{TextBackColor}{TextBackColor}{Input} \\
\\
\#\# Output Format \\
Given the user query, in your output, you first need to identify the user intent and the knowledge needed to solve the task in the user query. \\
Then, rate the difficulty level of the user query as \texttt{very easy}, \texttt{easy}, \texttt{medium}, \texttt{hard}, or \texttt{very hard}. \\
\\
Now, please output the user intent and difficulty level below in a json format by filling in the placeholders in []: \\
\{\{ \\
    ``intent'': ``The user wants to [....]'', \\
    ``knowledge'': ``To solve this problem, the models need to know [....]'', \\
    ``difficulty'': ``[very easy/easy/medium/hard/very hard]'' \\
\}\} \\
\end{tcolorbox}
\caption{The prompts used to judge the difficulty level of questions.}
\label{fig:prompt_difficulty_check}
\end{figure*}

\begin{figure*}
\begin{tcolorbox}[
  title=Prompts for Topic Classification,
  colback=white,
  colframe=black!70,
  fontupper=\ttfamily,
  coltitle=white,
  colbacktitle=blue!50!black,
  fonttitle=\ttfamily,
  boxrule=0.8pt,
]
As a mathematics education specialist, please analyze the topics of the provided question and its answer.\\
Specific requirements are as follows: \\
1. You should identify and categorize the main mathematical topics involved in the problem. If knowledge from non-mathematical fields is used, it is classified into Others - xxx, such as Others - Problem Context. \\
2. You should put your final answer between $<$TOPIC$>$ and $<$/TOPIC$>$. \\
------ \\
Question: Compute $\cos 330^\circ$. \\
\\
Answer: We know that $330^\circ = 360^\circ - 30^\circ$. \\
Since $\cos(360^\circ - \theta) = \cos \theta$ for all angles $\theta$, \\
we have $\cos 330^\circ = \cos 30^\circ$. \\
Since $\cos 30^\circ = \frac{\sqrt{3}}{2}$, \\
we can conclude that $\cos 330^\circ = \boxed{\frac{\sqrt{3}}{2}}$. \\
\\
Analysis: $<$TOPIC$>$Trigonometry - Cosine Function$<$/TOPIC$>$ \\
------ \\
Question: \fcolorbox{TextBackColor}{TextBackColor}{Question} \\
\\
Answer: \fcolorbox{TextBackColor}{TextBackColor}{Answer} \\
\\
Analysis:
\end{tcolorbox}
\caption{The prompts used for topic classification.}
\label{fig:prompt_topic}
\end{figure*}

\begin{figure*}[!t]
\begin{tcolorbox}[
  title=Examples for Solvability Optimization,
  colback=white,
  colframe=gray!60!black,
  fontupper=\ttfamily,
  coltitle=white,
  colbacktitle=green!30!black,
  fonttitle=\ttfamily,
  boxrule=0.8pt
]
\textbf{Problems 1 (Before Optimization):} \\
There are 10 survivors in an emergency room. Each survivor is either a child, a woman, or a man. If there are 4 men and \textcolor{ErrorRed}{3 times as many women as men}, how many children are there? \\
\textbf{Problems 1 (After Optimization):} \\
There are 10 survivors in an emergency room. Each survivor is either a child, a woman, or a man. If there are 4 men and \textcolor{CorrectGreen}{an equal number of women as men}, how many children are there?
\vspace{3pt} \hrule \vspace{3pt}
\textbf{Problems 2 (Before Optimization):} \\
How many sides does a polygon have \textcolor{ErrorRed}{if it is a regular polygon?} \\
\textbf{Problems 2 (After Optimization):} \\
How many sides does a regular polygon have \textcolor{CorrectGreen}{if each interior angle is 120 degrees}? \\
\vspace{-5pt} \hrule \vspace{3pt}
\textbf{Problems 3 (Before Optimization):} \\
Find the sum of the first three terms of \textcolor{ErrorRed}{this series}. \\
\textbf{Problems 3 (After Optimization):} \\
Calculate the sum of the first three terms of the \textcolor{CorrectGreen}{arithmetic series where the first term is 5 and the common difference is 3}.
\end{tcolorbox}
\caption{Three examples for solvability optimization by GPT-4o-mini.}
\label{fig:solvability_optim_example}
\end{figure*}

\begin{figure*}[!t]
\begin{tcolorbox}[
  title=Examples for Difficulty Optimization,
  colback=white,
  colframe=gray!60!black,
  fontupper=\ttfamily,
  coltitle=white,
  colbacktitle=green!30!black,
  fonttitle=\ttfamily,
  boxrule=0.8pt
]
\textbf{Problems 1 (Before Optimization):} \\
How many 4-digit positive integers are there? \\
\textbf{Problems 1 (After Optimization):} \\
How many 4-digit positive integers can be formed using non-repeating digits where the sum of these digits must be even, and the integers fall within the range of 1000 to 9999? \\
\vspace{-5pt} \hrule \vspace{3pt}
\textbf{Problems 2 (Before Optimization):} \\
The average of 15 numbers is 32. An additional number is then added to the list, and the new average of the 16 numbers is 34. What number was added to the list? \\
\textbf{Problems 2 (After Optimization):} \\
The average of 15 positive integers is 32, but one integer fluctuates to 30 before adding a new number. After adding this new number, the average of the 16 integers becomes 34. Calculate the added number and find the standard deviation of all 16 integers, considering their ascending order. \\
\vspace{-5pt} \hrule \vspace{3pt}
\textbf{Problems 3 (Before Optimization):} \\
A fair coin is tossed 50 times, what is the probability of getting heads at least 25 times? \\
\textbf{Problems 3 (After Optimization):} \\
A fair coin is tossed 50 times; what is the probability of obtaining heads at least 25 times, and can you also calculate the expected number, variance, and standard deviation of heads while determining the likelihood that the total number of heads exceeds 30?
\end{tcolorbox}
\caption{Three examples for difficulty optimization by GPT-4o-mini.}
\label{fig:difficulty_optim_example}
\end{figure*}

\begin{figure*}[!t]
\begin{tcolorbox}[
  title=ScaleQuest Data Example 1,
  colback=white,
  colframe=gray!60!black,
  fontupper=\ttfamily,
  coltitle=white,
  colbacktitle=green!30!black,
  fonttitle=\ttfamily,
  boxrule=0.8pt
]
\small
\textbf{QUESTION:} \\
My orthogonal projection matrix \( P \) projects onto the plane with equation \[x+y+z=0.\]Find the matrix \( P \). \\
\vspace{-5pt} \hrule \vspace{4pt}
\textbf{SOLUTION:} \\
To find the orthogonal projection matrix \( P \) that projects onto the plane with the equation \( x + y + z = 0 \), we can follow these steps:\\
1. \textbf{Identify the normal vector to the plane:}\\
The plane \( x + y + z = 0 \) has a normal vector \( \mathbf{n} = \begin{pmatrix} 1 \\ 1 \\ 1 \end{pmatrix} \).

2. \textbf{Express the projection matrix in terms of the normal vector:}\\
The orthogonal projection matrix \( P \) onto the plane can be written as:
\[
P = I - \frac{\mathbf{n} \mathbf{n}^T}{\mathbf{n}^T \mathbf{n}}
\]
where \( I \) is the \( 3 \times 3 \) identity matrix, \( \mathbf{n} \) is the normal vector, and \( \mathbf{n}^T \) is the transpose of the normal vector.

3. \textbf{Calculate \( \mathbf{n}^T \mathbf{n} \):}
\[
\mathbf{n}^T \mathbf{n} = \begin{pmatrix} 1 & 1 & 1 \end{pmatrix} \begin{pmatrix} 1 \\ 1 \\ 1 \end{pmatrix} = 1^2 + 1^2 + 1^2 = 3
\]

4. \textbf{Calculate \( \mathbf{n} \mathbf{n}^T \):}
\[
\mathbf{n} \mathbf{n}^T = \begin{pmatrix} 1 \\ 1 \\ 1 \end{pmatrix} \begin{pmatrix} 1 & 1 & 1 \end{pmatrix} = \begin{pmatrix} 1 & 1 & 1 \\ 1 & 1 & 1 \\ 1 & 1 & 1 \end{pmatrix}
\]

5. \textbf{Substitute these into the projection matrix formula:}
\[
P = I - \frac{1}{3} \begin{pmatrix} 1 & 1 & 1 \\ 1 & 1 & 1 \\ 1 & 1 & 1 \end{pmatrix}
\]
where \( I \) is the \( 3 \times 3 \) identity matrix:
\[
I = \begin{pmatrix} 1 & 0 & 0 \\ 0 & 1 & 0 \\ 0 & 0 & 1 \end{pmatrix}
\]

6. \textbf{Perform the matrix subtraction:}
\[
P = \begin{pmatrix} 1 & 0 & 0 \\ 0 & 1 & 0 \\ 0 & 0 & 1 \end{pmatrix} - \frac{1}{3} \begin{pmatrix} 1 & 1 & 1 \\ 1 & 1 & 1 \\ 1 & 1 & 1 \end{pmatrix}
\]
\[
P = \begin{pmatrix} \frac{2}{3} & -\frac{1}{3} & -\frac{1}{3} \\ -\frac{1}{3} & \frac{2}{3} & -\frac{1}{3} \\ -\frac{1}{3} & -\frac{1}{3} & \frac{2}{3} \end{pmatrix}
\]

Therefore, the orthogonal projection matrix \( P \) is:
\[
\boxed{\begin{pmatrix} \frac{2}{3} & -\frac{1}{3} & -\frac{1}{3} \\ -\frac{1}{3} & \frac{2}{3} & -\frac{1}{3} \\ -\frac{1}{3} & -\frac{1}{3} & \frac{2}{3} \end{pmatrix}}
\]
\end{tcolorbox}
\caption{An example from our constructed dataset.}
\label{fig:scalequest_example1}
\end{figure*}

\begin{figure*}[!t]
\begin{tcolorbox}[
  title=ScaleQuest Data Example 2,
  colback=white,
  colframe=gray!60!black,
  fontupper=\ttfamily,
  coltitle=white,
  colbacktitle=green!30!black,
  fonttitle=\ttfamily,
  boxrule=0.8pt
]
\small
\textbf{QUESTION:} \\
Suppose $a,b$ are positive real numbers such that $a+b = 1$.

(a) minimize $\frac{1}{a} + \frac{1}{b}$;

(b) minimize $a^2 + b^2$. \\
\vspace{-5pt} \hrule \vspace{4pt}
\textbf{SOLUTION:} \\
To solve these problems, we will use the method of Lagrange multipliers and the AM-HM inequality.

\textbf{Part (a):} Minimize $\frac{1}{a} + \frac{1}{b}$ given $a + b = 1$

First, we use the method of Lagrange multipliers. Define the function $f(a, b) = \frac{1}{a} + \frac{1}{b}$ and the constraint $g(a, b) = a + b - 1 = 0$.

The Lagrangian is:
\[
\mathcal{L}(a, b, \lambda) = \frac{1}{a} + \frac{1}{b} + \lambda (a + b - 1)
\]

Taking the partial derivatives and setting them to zero, we get:
\[
\frac{\partial \mathcal{L}}{\partial a} = -\frac{1}{a^2} + \lambda = 0 \implies \lambda = \frac{1}{a^2}
\]
\[
\frac{\partial \mathcal{L}}{\partial b} = -\frac{1}{b^2} + \lambda = 0 \implies \lambda = \frac{1}{b^2}
\]
\[
\frac{\partial \mathcal{L}}{\partial \lambda} = a + b - 1 = 0
\]

Since $\lambda = \frac{1}{a^2}$ and $\lambda = \frac{1}{b^2}$, we have:
\[
\frac{1}{a^2} = \frac{1}{b^2} \implies a^2 = b^2 \implies a = b \quad (\text{since } a, b > 0)
\]

Substituting $a = b$ into the constraint $a + b = 1$, we get:
\[
a + a = 1 \implies 2a = 1 \implies a = \frac{1}{2} \implies b = \frac{1}{2}
\]

Now, we calculate $\frac{1}{a} + \frac{1}{b}$:
\[
\frac{1}{a} + \frac{1}{b} = \frac{1}{\frac{1}{2}} + \frac{1}{\frac{1}{2}} = 2 + 2 = 4
\]

Thus, the minimum value of $\frac{1}{a} + \frac{1}{b}$ is $\boxed{4}$.

\textbf{Part (b):} Minimize $a^2 + b^2$ given $a + b = 1$

We use the identity for the square of a sum:
\[
(a + b)^2 = a^2 + b^2 + 2ab
\]

Given $a + b = 1$, we have:
\[
1^2 = a^2 + b^2 + 2ab \implies 1 = a^2 + b^2 + 2ab \implies a^2 + b^2 = 1 - 2ab
\]

To minimize $a^2 + b^2$, we need to maximize $ab$. Using the AM-GM inequality:
\[
\frac{a + b}{2} \geq \sqrt{ab} \implies \frac{1}{2} \geq \sqrt{ab} \implies \left(\frac{1}{2}\right)^2 \geq ab \implies \frac{1}{4} \geq ab
\]

The maximum value of $ab$ is $\frac{1}{4}$, which occurs when $a = b = \frac{1}{2}$.

Substituting $ab = \frac{1}{4}$ into the expression for $a^2 + b^2$:
\[
a^2 + b^2 = 1 - 2ab = 1 - 2 \left(\frac{1}{4}\right) = 1 - \frac{1}{2} = \frac{1}{2}
\]

Thus, the minimum value of $a^2 + b^2$ is $\boxed{\frac{1}{2}}$.
\end{tcolorbox}
\caption{An example from our constructed dataset.}
\label{fig:scalequest_example2}
\end{figure*}

\begin{figure*}[!t]
\begin{tcolorbox}[
  title=ScaleQuest Data Example 3,
  colback=white,
  colframe=gray!60!black,
  fontupper=\ttfamily,
  coltitle=white,
  colbacktitle=green!30!black,
  fonttitle=\ttfamily,
  boxrule=0.8pt
]
\small
\textbf{QUESTION:} \\
Compute
\[
\sum_{n = 1}^{\infty} \frac{4^n + 5^n + 6^n}{12^n}.
\]
\vspace{-5pt} \hrule \vspace{4pt}
\textbf{SOLUTION:} \\
To compute the sum $\sum_{n=1}^{\infty} \frac{4^n + 5^n + 6^n}{12^n}$, we can split the sum into three separate sums:

\[
\sum_{n=1}^{\infty} \frac{4^n + 5^n + 6^n}{12^n} = \sum_{n=1}^{\infty} \frac{4^n}{12^n} + \sum_{n=1}^{\infty} \frac{5^n}{12^n} + \sum_{n=1}^{\infty} \frac{6^n}{12^n}.
\]

Each of these sums is a geometric series. Let's evaluate each one separately.

1. For the first sum $\sum_{n=1}^{\infty} \frac{4^n}{12^n}$:
\[
\sum_{n=1}^{\infty} \frac{4^n}{12^n} = \sum_{n=1}^{\infty} \left(\frac{4}{12}\right)^n = \sum_{n=1}^{\infty} \left(\frac{1}{3}\right)^n.
\]
The sum of an infinite geometric series $\sum_{n=1}^{\infty} ar^{n-1}$ is given by $\frac{a}{1-r}$ for $|r| < 1$. Here, $a = \frac{1}{3}$ and $r = \frac{1}{3}$, so:
\[
\sum_{n=1}^{\infty} \left(\frac{1}{3}\right)^n = \frac{\frac{1}{3}}{1 - \frac{1}{3}} = \frac{\frac{1}{3}}{\frac{2}{3}} = \frac{1}{2}.
\]

2. For the second sum $\sum_{n=1}^{\infty} \frac{5^n}{12^n}$:
\[
\sum_{n=1}^{\infty} \frac{5^n}{12^n} = \sum_{n=1}^{\infty} \left(\frac{5}{12}\right)^n.
\]
Here, $a = \frac{5}{12}$ and $r = \frac{5}{12}$, so:
\[
\sum_{n=1}^{\infty} \left(\frac{5}{12}\right)^n = \frac{\frac{5}{12}}{1 - \frac{5}{12}} = \frac{\frac{5}{12}}{\frac{7}{12}} = \frac{5}{7}.
\]

3. For the third sum $\sum_{n=1}^{\infty} \frac{6^n}{12^n}$:
\[
\sum_{n=1}^{\infty} \frac{6^n}{12^n} = \sum_{n=1}^{\infty} \left(\frac{6}{12}\right)^n = \sum_{n=1}^{\infty} \left(\frac{1}{2}\right)^n.
\]
Here, $a = \frac{1}{2}$ and $r = \frac{1}{2}$, so:
\[
\sum_{n=1}^{\infty} \left(\frac{1}{2}\right)^n = \frac{\frac{1}{2}}{1 - \frac{1}{2}} = \frac{\frac{1}{2}}{\frac{1}{2}} = 1.
\]

Adding these three results together, we get:
\[
\sum_{n=1}^{\infty} \frac{4^n + 5^n + 6^n}{12^n} = \frac{1}{2} + \frac{5}{7} + 1.
\]

To add these fractions, we need a common denominator. The least common multiple of 2, 7, and 1 is 14. So we convert each fraction:
\[
\frac{1}{2} = \frac{7}{14}, \quad \frac{5}{7} = \frac{10}{14}, \quad 1 = \frac{14}{14}.
\]

Adding these fractions together, we get:
\[
\frac{7}{14} + \frac{10}{14} + \frac{14}{14} = \frac{7 + 10 + 14}{14} = \frac{31}{14}.
\]

Thus, the sum is:
\[
\boxed{\frac{31}{14}}.
\]
\end{tcolorbox}
\caption{An example from our constructed dataset.}
\label{fig:scalequest_example3}
\end{figure*}

\end{document}